# Locomotion Dynamics of an Underactuated Three-Link Robotic Vehicle

Leonid Raz (Rizyaev) and Yizhar Or

***Abstract*— The wheeled three-link snake robot is a well-known example of an underactuated system modelled using nonholonomic constraints, preventing lateral slippage (skid) of the wheels. A kinematically controlled configuration assumes that both joint angles are directly prescribed as phase-shifted periodic input. In another configuration of the robot, only one joint is periodically actuated while the second joint is passively governed by a visco-elastic torsion spring. In our work, we constructed the two configurations of the wheeled robot and conducted motion experiments under different actuation inputs. Analysis of the motion tracking measurements reveals a significant amount of wheels' skid, in contrast to the assumptions used in standard nonholonomic models. Therefore, we propose modified dynamic models which include wheels' skid and viscous friction forces, as well as rolling resistance. After parameter fitting, these dynamic models reach good agreement with the motion measurements, including effects of input's frequency on the mean speed and net displacement per period. This illustrates the importance of incorporating wheels' skid and friction into the system's model.**

## I. INTRODUCTION

THIS work studies the locomotion dynamics of an underactuated wheeled three-link snake. Let us first explain some key terms regarding the research topic. ***Locomotion*** is a general term for movement or the ability to move from one place to another. Within the context of this study, the more specific term is undulatory locomotion, which refers to a specific type of movement characterized by time-periodic input of shape actuation, which generates motion. An example of this could be the walking or swimming of humans or crawling movement of snakes. The periodic shape change is called a gait. The gait dictates the timing, amplitude, and phase relationships of these movements.

The movement of snake-like robots has been extensively studied and provides a fascinating bio-inspired example. It involves various mechanical factors, such as efficiency, stability, and the capability to traverse different types of terrain.

It is worth mentioning the significant contributions of Shigeo Hirose [1] in the field of biologically inspired hyper-redundant robots. Hirose explored the mechanics of snake-like movement and conducted many experiments with robotic systems that mimic snakes.

He investigated the various gaits in both two and three dimensions, using robots with multiple linked segments. These experiments were carried out on planar surfaces as well as on rough terrain and in aquatic environments. The robots' redundancy allows them to continue functioning even if certain components or segments are damaged or malfunctioning, enhancing their reliability and performance in critical operations, such as search and rescue missions [2], [3]. To effectively accomplish these tasks, it is essential to employ motion planning methods as researched in [4].

The next concept to be explained is ***underactuation***. Underactuated robots are mechanical systems in which not all degrees of freedom are actively controlled or actuated. They are found in fields like robotics, automation, and biomechanics. This is indeed a comprehensive definition that includes both living organisms, such as humans and animals, as well as mechanical systems like cars and airplanes. Consider a mechanical system with $n$ degrees of freedom, where the locations of its components can be described by a vector of generalized coordinates $\mathbf{q} = (\mathrm{q}_1, q_2, \dots q_n)^T$. In practice, these systems often have motion-limiting constraints. Constraints which impose restrictions of the possible motions of the components and can be described solely in terms of the system's coordinates, are called ***holonomic constraints***. Constraints of this type can be formulated as scalar equations of the form $h_i(q_1, \dots, q_n) = 0$ for $i = (1,2, \dots m)$, or in vector form $\mathbf{h}(\mathbf{q}) = 0$. Another type of constraints are ***nonholonomic constraints*** [5], [6], and they impose restrictions on the possible directions of the system's generalized velocities $\dot{\mathbf{q}}$. Constraints of this type can be written in the form of $\mathbf{f}(\mathbf{q}, \dot{\mathbf{q}}) = \mathbf{0}$, where $\mathbf{q}$ is the vector of generalized coordinates. The nonholonomic constraints are typically nonintegrable, meaning they cannot be formulated as constraints on the coordinates only. Nonholonomic constraints can be observed in various fields. In space applications, satellites serve as an example of this as they control their spatial orientation by using flywheels, relying on the principle of angular momentum conservation, which is considered nonholonomic [7]. Another earthly example of a purely kinematic, nonholonomic system is the Dubins' car [8]. It can represent all types of vehicles with wheels that cannot slip laterally, parallel to their own axis of rotation, which imposes restrictions on the possible paths that the vehicle can travel. Lateral slippage will be referred to as *skid* in this work from here on. Another classical examples of a nonholonomic system is the Chaplygin's sleigh [9], [10], a rigid body sliding on a planar surface, with a fixed blade which prevents the skid of a

This work was supported in part by Israel Science Foundation under grant no. 1382/23, and Israel Ministry of Innovation, Science and Technology under grant no. 3-17383, and Kahn Foundation grant no. 2029753 under Technion's Autonomous Systems program.

L. Raz (Rizyaev) and Y. Or are with the Faculty of Mechanical Engineering, Technion – Israel Institute of Technology, Haifa 3200003, Israel. *(Corresponding author: Yizhar Or, email: izi@technion.ac.il).*

given body point along a body-fixed direction. Similarly, a rolling disk [11] whose contact point with the ground is constrained not to slip and skid. These examples can be generalized to all wheeled vehicles that move on surfaces with high friction. Underactuated robotic vehicles with nonholonomic constraints can be broadly categorized into two types: those that employ direct actuation of wheels' rotation, such as bicycles and cars, and those where all wheels are rotating passively, and the actuation is applied at articulated joints. Examples are the Twistcar [12] and the Snakeboard [13], [14]. A further distinction can be made between kinematic and dynamic systems. Assume a system with $n$ degrees of freedom, $k$ nonholonomic constraints, and $m$ kinematically prescribed inputs. If $k = n - m$ holds, the system is kinematic. That is, its motion is governed by first-order differential equations generated by the nonholonomic constraints. On the other hand, if $k < n - m$, that is there are less constraints than the unactuated degrees of freedom, the system is dynamic, and its motion is governed by a combination of nonholonomic constraints and second-order differential equations representing momentum evolution.

Another category of systems with articulated joints that share common characteristics are the multi-link swimmers. Similar to a blade or a wheel that prevents sideways skidding, the non-holonomic constraints of the swimmers mainly arise from the interaction between the links and the surrounding fluid. Two main types of swimmers can be distinguished. Micro-swimmers, whose movement is governed primarily by fluid viscosity, derive their nonholonomic constraints from the quasi-static equilibrium within the fluid, while inertial effects are negligible [15], [16]. On the other hand, the motion of larger swimmers relies on inertia, resulting in nonholonomic constraints that originate from the conservation of momentum, while neglecting viscous drag effects [17], [18].

In addition to the cases mentioned above, in which there is full kinematic control over the shape variables or joints, there is another category of models that involve passive joints, such as a passive tail of an articulated swimmer [19], [20], or the passive joints of a snake robot [21]. These models incorporate joints that are not actively controlled or actuated, and their motion is determined by visco-elasticity, coupled with external forces and interactions with the environment. Since these cases are not purely kinematic systems, they require formulation of the system's complete dynamic equations.

In real-world scenarios of wheeled robots, assuming no-skid nonholonomic constraints is often unrealistic [22]. Therefore, modified models can be developed that consider skidding and other energy dissipation mechanisms. By accounting for energy dissipation, these modified models have the potential to yield more realistic results. Three main approaches of relaxing the no-skid nonholonomic constraints can be identified in previous research. Some studies suggest the skid angle model [23], [24], [25], [26]. This model adjusts the nonholonomic constraints, which limit the possible movements of the system, by rotating them by a time-varying angle called "skid angle". By doing so, the model can incorporate the skidding effect while preserving the structure of nonholonomic constraints. Nontheless, the skid angles for each wheel are additional degrees-of-freedom, which increases the dimensionality of the system's dynamics. Furthermore, this model does not introduce energy dissipation, which is a key effect in wheels skid. Another approach, which has already been mentioned, is the Coulomb-type friction model [27], [28], [29], [30]. Simply put, this model states that the friction force is directly related to the normal force and is limited by it. This leads to discontinuity of the skidding velocity and reaction force when the skid velocity crosses zero, i.e. stick-slip transitions. This friction model includes energy dissipation and requires the use of hybrid dynamics to transition between the different states of the system [29]. A main disadvantage of this friction model is that it leads to non-smooth dynamical system with multiple states and transitions, which causes severe issues of high sensitivity in the numeric integration process. Finally, the third approach, which we chose to use in this work, is viscous resistance model, in which the friction force is assumed to be linearly proportional to wheel's velocity. This model combines the advantages of a continuous and smooth model together with consideration of energy dissipation [31], [32], [33]. In this model, the frictional forces have components that are linearly proportional to both skid and roll velocities. It will serve as the primary frictional model for considering all dissipation mechanisms in this study.

The wheeled three-link snake [22], which is the basic model of this research is also a well-known example of an underactuated system which is modelled using nonholonomic constraints, preventing skidding of the wheels. A kinematically controlled version assumes that both joint angles $(\phi_1, \phi_2)$ are directly prescribed as phase-shifted periodic input, see **Fig. 1** and **Fig. 2**. The kinematic three-wheel snake was treated in [22], with emphasis on gait planning. Skidding was observed in their motion experiments, and it was mentioned that it was of a greater magnitude when the gait crosses symmetric configurations where the nonholonomic constraints become singular. Quantitative measurements were not reported, and the skidding was not accounted for in their numerical simulations. In addition, the effect of actuation frequency was not studied in [22].

A later work [29] studied the dynamics of the kinematically controlled three-wheel snake robot near its singular symmetric states, using numerical simulations. This work accounted for Coulomb-type friction and incorporated hybrid dynamics to analyze the robot's dynamics while crossing singularity. As shown in [29], Coulomb-type friction leads to stick-slip transitions, non-smoothness, and numerical sensitivity. This work investigated the effect of both frequency and amplitude on the robot's motion using numerical simulations, but no experiments were performed.

In the semi-passive version of the three-wheel snake robot, only one joint is periodically actuated, while the second joint is passively governed by a visco-elastic torsion spring. Some previous works have studied this type of robot [34] and even a multi-link version [21], with nonholonomic constraints. In [34] numerical simulations of a semi-passive model, which crosses singularity, have been performed. In [21] motion tracking experiments were presented for three-link and four-link wheeled robots with passive-elastic joints and a single actuated joint. However, no specific measurement was provided in [21] for skidding, and there was no proposed theoretical model that considers skidding and energy dissipation.

The goal of this research is to extend previous studies of the wheeled three-link snake robot, both theoretically and experimentally. A dedicated robot was constructed for the research, as depicted in **Fig. 1**, for both configurations of two kinematically-actuated joints (**Fig. 2**(a)) and semi-passive case (**Fig. 2**(b)). We generalize the theoretical models to incorporate the effect of wheels' skid and frictional rolling resistance on the robot's dynamic motion. In addition, we aim at conducting systematic motion tracking experiments to verify the generalized theoretical models. Next, we perform a parameter fitting process to calibrate the models' parameters and compare them with the experimental results, as well as assessing the suitability of different dissipation models to the various scenarios tested. Finally, we study the significant effect of actuation frequency on the motion, in the two actuation configurations, both in simulations and experiments. For brevity of the presentation, some extensions of the theoretical analysis and further details on the experiments are relegated to [35].

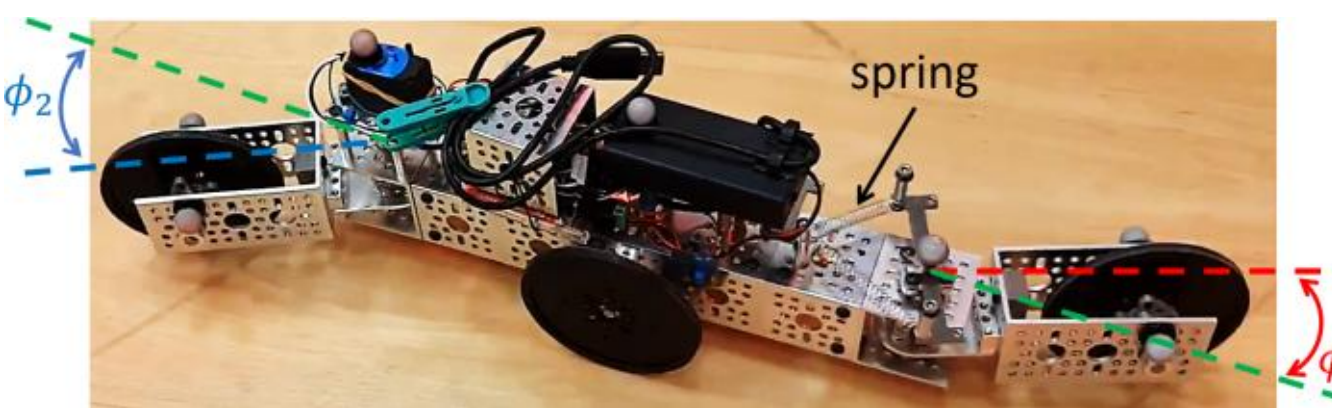

**Fig. 1**: The three-link wheeled robot, constructed as a part of this research, in its semi-passive congifuration. The spring is installed for a symmetric gait input.

## II. THEORETICAL ANALYSIS

This section presents the basic model of this research – a wheeled three-link snake robot. It is divided into two main configurations: kinematic shape-actuation (**Fig. 2**(a)) and semi-passive shape-actuation (**Fig. 2**(b)). The model and its assumptions are discussed first, followed by the formulation of equations of motion using the Lagrange method. Finally, the formulation is extended to incorporate energy dissipation, thus generalizing the models and making them more realistic.

### *A. Kinematic Model*

The planar model of the robot, shown in **Fig. 2**(a), consists of three rigid links connected by two rotary joints. The robot is supported by wheels. Ideally, the wheels are assumed to resist skidding, which translates into non-holonomic constraints on the robot's motion. The dimensions of the links are shown in the figure, and they are assumed to have a uniform mass distribution. Referring to **Fig. 2**, $h$ is half the length of the middle link, $b$ is the distance of the side link's center of mass from the joint, and $l$ is the distance of the side link's wheel from the joint.

We define vector $\mathbf{q} = (x, y, \theta, \phi_1, \phi_2)^T$ of generalized coordinates that describe the robot's motion. Accordingly, the robot is a system of five degrees of freedom. As seen in **Fig. 2**, $(x, y)$ describe the position of the center of link 0, and $\theta$ describes its orientation angle. Coordinates $(\phi_1, \phi_2)$ are the angles of the side links, measured relative to the middle link. At this point, angles $\phi_1, \phi_2$ are prescribed as periodic functions of time, i.e., the joints are actively controlled. The vector $\mathbf{q}$ can be divided into two parts $\mathbf{q} = (\mathbf{q}_\text{b}, \mathbf{q}_\text{s})^T$. A part which defines the position of the robot in the plane $\mathbf{q}_\text{b} = (x, y, \theta)^T$, and another part which defines its shape $\mathbf{q}_\text{s} = (\phi_1, \phi_2)^T$.

The robot has three wheels that are in contact with the ground. The wheels are assumed to resist skidding – thus, allowing only a rolling motion parallel to the longitudinal direction of each link. These no-skid constraints are essentially nonholonomic, since they impose restrictions on the velocities of the system. Mathematically, these constraints can be formulated as $v_i^\perp = \dot{\mathbf{r}}_i \cdot \hat{\mathbf{n}}_i = 0$ for $i = 0,1,2$. Where $\dot{\mathbf{r}}_i$ is the position vector of each wheel and $\hat{\mathbf{n}}_i$ is the perpendicular vector for each link (see **Fig. 2**).

The robot has five degrees of freedom, where two of them are inputs since they are actively controlled. The additional three equations required to solve the robot's motion are derived from the no-skid constraints. Written in matrix form (1):

(a)

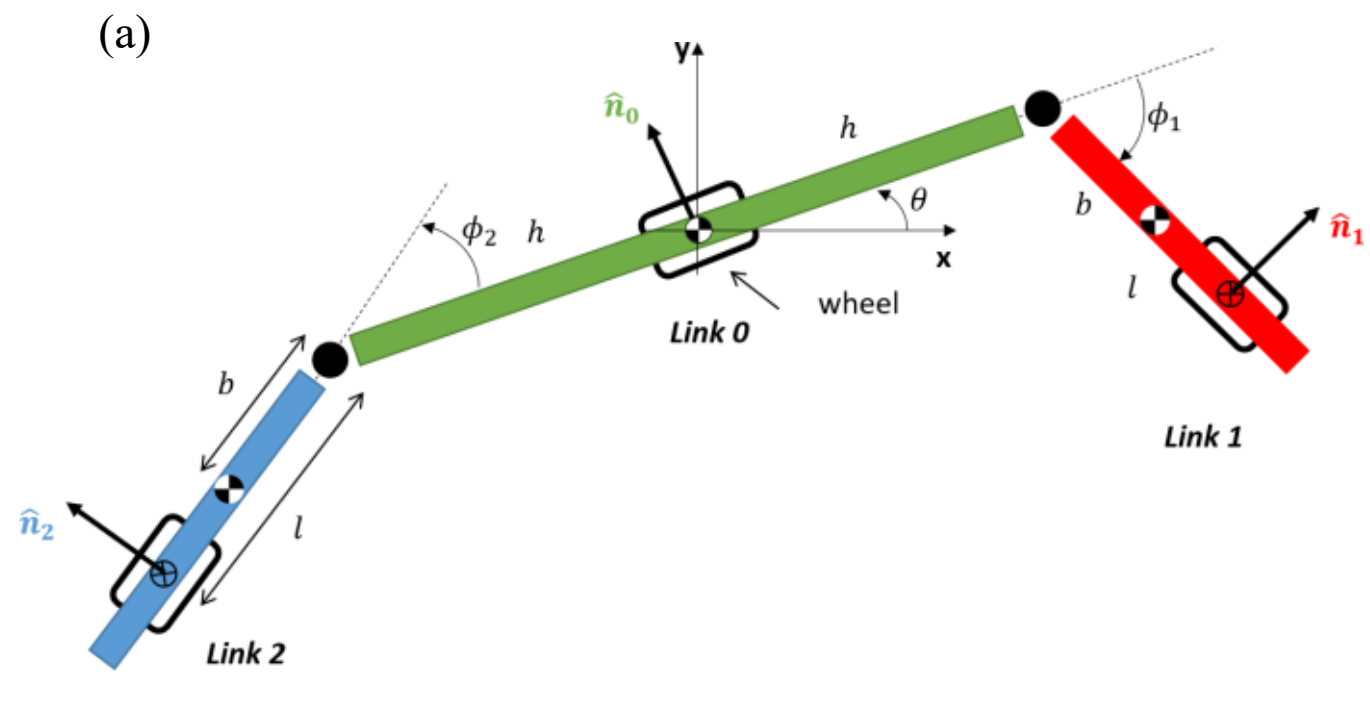

(b)

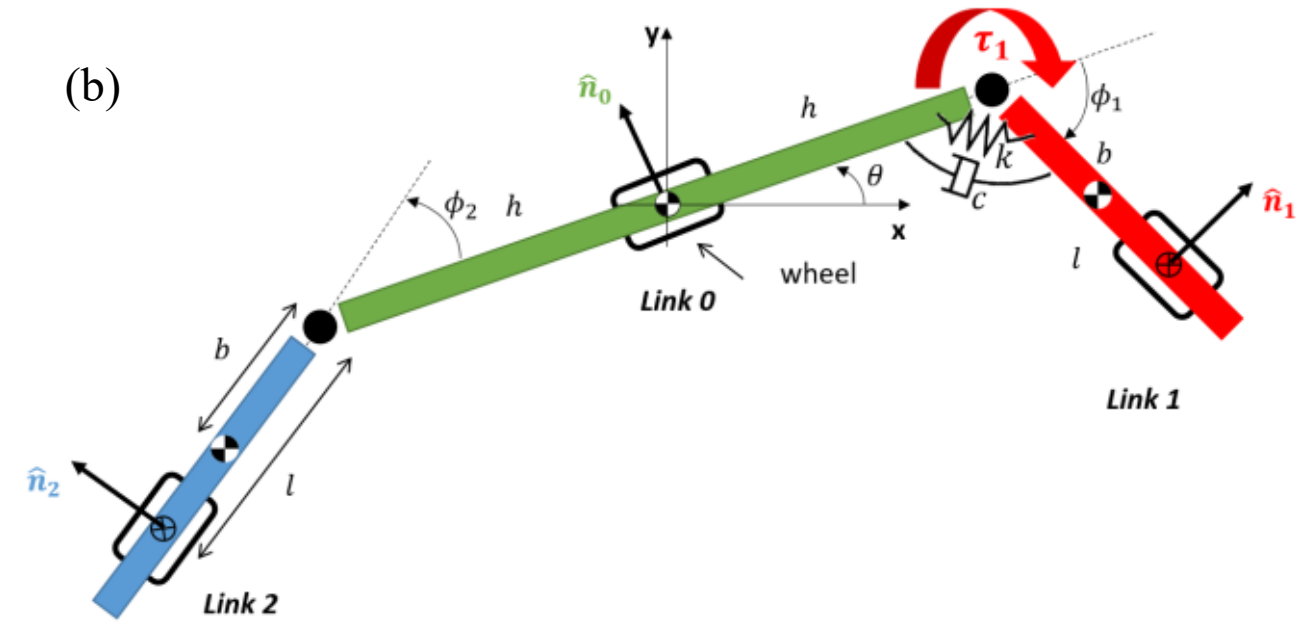

**Fig. 2.** The planar three-wheel snake robot models: (a) Kinematic shape-actuated model (b) Semi-passive model.

$$\mathbf{W}(\mathbf{q})\dot{\mathbf{q}} = 0$$

Where:

$$\mathbf{W}(\mathbf{q}) = \Bigg[\underbrace{\begin{matrix} -\sin(\theta) & \cos(\theta) & 0 \\ \sin(\phi_1 - \theta) & \cos(\phi_1 - \theta) & l + h\cos(\phi_1) \\ -\sin(\phi_2 + \theta) & \cos(\phi_2 + \theta) & -l - h\cos(\phi_2) \end{matrix}}_{\mathbf{W}_b(\mathbf{q})} \quad \underbrace{\begin{matrix} 0 & 0 \\ -l & 0 \\ 0 & -l \end{matrix}}_{\mathbf{W}_s(\mathbf{q})}\Bigg] \tag{1}$$

By using the body and shape decomposition, (1) can be written as:

$$\mathbf{W}(\mathbf{q})\dot{\mathbf{q}} = \mathbf{W}_b(\mathbf{q})\dot{\mathbf{q}}_b + \mathbf{W}_s(\mathbf{q})\dot{\mathbf{q}}_s = 0 \tag{2}$$

This set of equations can be rearranged to solve for the unknown variables, $\dot{\mathbf{q}}_b$:

$$\dot{\mathbf{q}}_b = -\mathbf{W}_b^{-1}(\mathbf{q})\mathbf{W}_s(\mathbf{q})\dot{\mathbf{q}}_s \tag{3}$$

Equation (3) is a system of 1st-order differential equations. That is, the motion of the body is directly dictated by prescribing the shape variables, so that $\mathbf{q}_s(t)$ is a prescribed input. The solution will yield $\mathbf{q}_b(t)$, thus obtaining the system's body motion. It is important to note that this holds as long as matrix $\mathbf{W}_b$ in (3) is invertible, mathematically requiring that $det(\mathbf{W}_b) \neq 0$. From (1) we can calculate the determinant of $\mathbf{W}_b$:

$$\det(\mathbf{W}_b) = h \cdot sin(\phi_2 - \phi_1) + l(sin\phi_2 - sin\phi_1) \tag{4}$$

From (4) we can find two types of singular states where $det(\mathbf{W}_b) = 0$, which are thoroughly discussed in previous research [29]. Examining (3), when the model approaches singularity, the body velocities $\dot{\mathbf{q}}_b$ can grow unbounded. However, there is an exception – where while crossing a singularity, the body velocities remain bounded. This condition appears in (5). This case is also analyzed and explained in detail in previous work [29].

$$\dot{\phi}_1 \sin(\phi_2) + \dot{\phi}_2 \sin(\phi_1) = 0 \tag{5}$$

Considering symmetric singular configurations, where $\phi_1 = \phi_2$, (5) is reduced to (6) below. Stating that if the angular velocities follow this rule, unbounded divergence of $\dot{\mathbf{q}}_b$ is avoided [29]:

$$\dot{\phi}_1 = -\dot{\phi}_2 \tag{6}$$

*B. Formulation of the Dynamic Model*

The system's dynamic equations of motion are formulated using Lagrange's method, which is based on balance of mechanical work and potential energy [36], [37]. Casting into canonical matrix form, the system's dynamic equations are given as

$$\mathbf{M}(\mathbf{q})\ddot{\mathbf{q}} + \mathbf{B}(\mathbf{q}, \dot{\mathbf{q}}) + \mathbf{G}(\mathbf{q}) = \mathbf{E}\boldsymbol{\tau} + \mathbf{W}(\mathbf{q})^{\mathbf{T}}\boldsymbol{\Lambda} + \mathbf{F}_{\mathrm{d}} \tag{7}$$

$\mathbf{M}(\mathbf{q})$ is the system's inertia matrix and vector $\mathbf{B}(\mathbf{q}, \dot{\mathbf{q}})$ includes velocity-related terms. $\mathbf{G}(\mathbf{q}) = dU/d\mathbf{q}$ is the vector of generalized potential forces. On the right-hand side of (7), $\boldsymbol{\tau}$ is the vector of actuation torques at the joints, and $\mathbf{E}$ is a binary matrix such that $\mathbf{E}\boldsymbol{\tau} = [0 \;\; 0 \;\; 0 \;\; \tau_1 \;\; \tau_2]^T$. $\mathbf{W}(\mathbf{q})$ is the matrix of nonholonomic constraints, $\boldsymbol{\Lambda}$ is the vector of constraint forces and $\mathbf{F}_{\mathrm{d}} = -\partial D/\partial\dot{\mathbf{q}}$ is the vector of generalized dissipation forces, where $D(q, \dot{q})$ is Rayleigh dissipation function. The full derivation and expressions for $\mathbf{M}(\mathbf{q})$, $\mathbf{B}(\mathbf{q}, \dot{\mathbf{q}})$ for the three-wheel snake model appear in [35].

In order to generalize the formulation to cases where only part of the shape variables are directly prescribed, we update our terminology of *body* and *shape* coordinates to *passive* and *active* coordinates. That is, the vector of coordinates is decomposed as $\mathbf{q} = (\mathbf{q}_{\mathrm{p}}, \mathbf{q}_a)$. In the kinematic shape-actuated configuration, $\mathbf{q}_{\mathrm{p}} = \mathbf{q}_{\mathrm{b}}$ and $\mathbf{q}_a = \mathbf{q}_{\mathrm{s}}$. In the semi-passive case, $\mathbf{q}_{\mathrm{p}} = (x, y, \theta, \phi_1)$ and $\mathbf{q}_a = \phi_2$. By decomposing into blocks according to the passive and active coordinates, (7) can be formulated as

$$\begin{bmatrix} \mathbf{M}_{pp} & \mathbf{M}_{pa} \\ \mathbf{M}_{pa}^T & \mathbf{M}_{aa} \end{bmatrix} \begin{pmatrix} \ddot{\mathbf{q}}_p \\ \ddot{\mathbf{q}}_a \end{pmatrix} + \begin{pmatrix} \mathbf{B}_p \\ \mathbf{B}_a \end{pmatrix} + \begin{pmatrix} \mathbf{G}_p \\ \mathbf{G}_a \end{pmatrix} = \begin{pmatrix} \mathbf{0} \\ \boldsymbol{\tau} \end{pmatrix} + \begin{pmatrix} \mathbf{W}_{\mathrm{b}}^{\mathrm{T}} \\ \mathbf{W}_{\mathrm{s}}^T \end{pmatrix} \boldsymbol{\Lambda} + \begin{pmatrix} \mathbf{F}_{\mathrm{d}_p} \\ \mathbf{F}_{\mathrm{d}_a} \end{pmatrix} \tag{8}$$

For the kinematic shape-actuated configuration, one has $\boldsymbol{\tau} = (\tau_1, \tau_2)^T$, whereas for the semi-passive case, one has $\boldsymbol{\tau} = \tau_2$. To complete this set of equations, we can again use the nonholonomic constraint in (2) and differentiate it with respect to time to obtain:

$$\begin{aligned} \mathbf{W}(\mathbf{q})\ddot{\mathbf{q}} + \dot{\mathbf{W}}(\mathbf{q}, \dot{\mathbf{q}})\dot{\mathbf{q}} & \\ = \mathbf{W}_p(\mathbf{q})\ddot{\mathbf{q}}_p + \dot{\mathbf{W}}_p(\mathbf{q}, \dot{\mathbf{q}})\dot{\mathbf{q}}_p & \\ + \mathbf{W}_a(\mathbf{q})\ddot{\mathbf{q}}_a + \dot{\mathbf{W}}_a(\mathbf{q}, \dot{\mathbf{q}})\dot{\mathbf{q}}_a = 0 & \end{aligned} \tag{9}$$

We can recall that in the kinematic case, the vector $\mathbf{q}_{\mathrm{s}}(t)$ and its time derivatives are directly prescribed. Therefore, vector $\boldsymbol{\tau}$, the vector of torques enforcing it, is unknown. After solving the body coordinates vector $\mathbf{q}_b$ in (3), we can complete the dynamic analysis of the system and find the body accelerations $\ddot{\mathbf{q}}_b$, actuation torques $\boldsymbol{\tau}$, and constraint forces $\boldsymbol{\Lambda}$. It is worth emphasizing that since we are still dealing with a kinematic system, this set of equations is solved algebraically. That is, by combining (8) and (9) we get a set of linear differential-algebraic equations shown in (10) below.

$$\begin{aligned} & \underbrace{\begin{bmatrix} \mathbf{M}_{pp} & -\mathbf{W}_p^T \\ \mathbf{W}_p & \mathbf{0}_{3\times3} \end{bmatrix}}_{\mathbf{A}} \begin{pmatrix} \ddot{\mathbf{q}}_p \\ \boldsymbol{\Lambda} \end{pmatrix} \\ & = -\begin{pmatrix} \mathbf{M}_{pa}\ddot{\mathbf{q}}_a + \mathbf{B}_p + \mathbf{G}_p - \mathbf{F}_{\mathrm{d}_p} \\ \mathbf{W}_a(\mathbf{q})\ddot{\mathbf{q}}_a + \dot{\mathbf{W}}_p(\mathbf{q}, \dot{\mathbf{q}})\dot{\mathbf{q}}_p + \dot{\mathbf{W}}_a(\mathbf{q}, \dot{\mathbf{q}})\dot{\mathbf{q}}_a \end{pmatrix} \end{aligned} \tag{10}$$

In order to obtain a solution of (10), matrix $\mathbf{A}$ must be invertible. For the kinematic shape-actuated configuration, one obtains that $\det(\mathbf{A}) = -\det(\mathbf{W}_b)^2$ [29]. That is, the dynamic solution of (10) becomes ill-defined when crossing singular states as dictated by (4). As shown in [29], even in the singular case where the body velocities remain bounded (that is, the condition mentioned in (6) is satisfied), the constraint forces and actuation torques are not bounded. Therefore, one has to incorporate wheels' skid into our model, as will be discussed in the final part of this section.

*C. Semi-Passive Configuration*

We now consider the semi-passive configuration by modifying the model and introducing a compliant joint. We address the mathematical modeling of the passive joint and the subsequent updates to the Lagrange equations. The previously discussed model had two directly prescribed joints. Now, one joint has been replaced by a spring-damper mechanism, thus making it a passive joint, as shown in **Fig. 2**(b). The reaction torque $\tau_1$ acting at the passive joint is described in (11).

$$\tau_1 = -k_\tau(\phi_1(t) - \gamma_1) - c_\tau\dot{\phi}_1(t) \tag{11}$$

The torsional stiffness coefficient is $k_\tau$, $c_\tau$ is the torsional damping coefficient and $\gamma_1$ is the free position angle. The torsion spring introduces potential energy to the system which is expressed as

$$U(\mathbf{q}) = \frac{1}{2} \cdot k_\tau \cdot (\phi_1 - \gamma_1)^2 \tag{12}$$

Passive joint damping is added to the equations as part of the Rayleigh dissipation function since it dissipates energy from the overall system, formulated in (13) below.

$$D^C(\mathbf{q}, \dot{\mathbf{q}}) = \frac{1}{2} c\dot{\phi}_1^2(t) \tag{13}$$

Using the formulation in (8)-(10) with the apaptation of $\mathbf{q}_\mathrm{p}, \mathbf{q}_a$ to the semi-passive configuration as explained above, it can be shown that the matrix $\mathbf{A}$ in (10) satisfies $\det(\mathbf{A}) \neq 0$ for all physically-relevant values. That is, the dynamic equations for the semi-passive configuration are not singular, hence the generalized forces and accelerations remain bounded as long as the kinematic singularity is overcome, such that the body velocities are bounded. This is in contrast to the shape-actuated configuration.

### *D. Adding Wheels' Dissipative Resistance*

We now address the incorporation of skid and roll dissipation to the system. This is in addition to the damping of the compliant joint, which has been treated in (14). We consider two mechanisms of energy dissipation: rolling resistance and skidding resistance of the wheels. The wheels' perpendicular velocities $v_i^\perp = \dot{\mathbf{r}}_i \cdot \hat{\boldsymbol{n}}_i$ have already been used in the no-skid constraints (1). Similarly, the wheels' tangential velocities are defined as $v_i^\parallel = \dot{\mathbf{r}}_i \cdot \hat{\mathbf{t}}_i$ (such that $\hat{\mathbf{t}}_i \cdot \hat{\boldsymbol{n}}_i = 0$). The velocities in the tangent direction are referred to as roll velocities, while the velocities in the perpendicular direction are denoted as skid velocity.

The roll dissipation force at each wheel $i$ is expressed in (14), alongside the appropriate Rayleigh dissipation term $D_i^R$.

$$f_i^R(\dot{\mathbf{q}}) = -c_i^R v_i^\parallel \quad ; \quad D_i^R(\mathbf{q}, \dot{\mathbf{q}}) = \frac{1}{2} c_i^R {v_i^\parallel}^2 \tag{14}$$

The skid dissipation force at each wheel $i$ is expressed in (15), alongside the appropriate Rayleigh dissipation term $D_i^S$.

$$f_i^S(\dot{\mathbf{q}}) = -c_i^S v_i^\perp \quad ; \quad D_i^S(\mathbf{q}, \dot{\mathbf{q}}) = \frac{1}{2} c_i^S {v_i^\perp}^2 \tag{15}$$

The total Rayleigh dissipation function is the sum of those three contributors, as formulated in (16) below.

$$D(\mathbf{q}, \dot{\mathbf{q}}) = D^C + D^R + D^S \tag{16}$$

Having established the groundwork for various models and configurations, we now provide a comprehensive summary and organization of all the types. All the dynamic equations of motion are derived from (8) with the appropriate selections of $U$, $D$ and $\mathbf{W}$. For the kinematic shape-actuated configuration, without dissipation, set $U$=$D$=0. For the semi-passive configuration, employ $U$ as shown in (12) and introduce $D = D^C$ as shown in (13). To incorporate roll resistance to the models, use $D = D^R$ for the kinematic configuration and $D = D^C + D^R$ for the semi-passive. When adding skid resistance, nonholonomic constraints are disregarded by setting $\mathbf{W}$=0. For kinematic actuation use $D = D^R + D^S$, and for semi-passive use $D = D^C + D^R + D^S$.

## III. Numerical Simulations

In this section, we present results of numerical simulations and computational investigation of the theoretical dynamic models. Furthermore, the numerical simulations are also utilized for comparison with our experimental work in the next section. These simulations build upon the previously established theoretical basis and allow us to explore the dynamics of our system under various actuation inputs. All numerical simulations were performed in MATLAB, using the ODE45 solver. In each case, one needs to bring the equations of motion into the form of state equations $d\mathbf{z}/dt = \mathbf{f}(\mathbf{z}, t)$ and use ODE45 for numerical integration with the appropriate initial conditions.

We define the displacement per cycle, denoted as $d$, as the Euclidean distance between the center positions $\mathbf{r}_0 = [x, y]$ of the central link, at the beginning of two consecutive cycles. Consequently, the mean velocity, denoted as $\bar{v}$, is defined by dividing the displacement per cycle by the actuation period time. The expressions are presented in (17) below.

$$t_p = 2\pi/\omega \qquad d = \left|r_0(t_p) - r_0(0)\right| \qquad \bar{v} = d/t_p \tag{17}$$

### *A. Semi-Passive Configuration, No-Skid Dynamics*

We now present simulation results of the semi-passive configuration, assuming no-skid nonholonomic constraints. In this configuration, only one joint angle is directly prescribed. In our case this angle is $\phi_2$, as shown in (18) below.

$$\phi_2(t) = \alpha_2 \cos(\omega t) \tag{18}$$

The passive joint's angle is governed by the torque generated by the visco-elastic spring, as given in (11). The load-free position angle $\gamma_1$ is determined according to the desired actuation input, with $\gamma_1$ set to 0° for a symmetric case. The state vector $\mathbf{z}$ is constructed as follows: $\mathbf{z} = \left[\mathbf{q}_p, \dot{\mathbf{q}}_p\right]^T = \left[x, y, \theta, \phi_1, \dot{x}, \dot{y}, \dot{\theta}, \dot{\phi}_1\right]^T$. The solution for $\boldsymbol{z}(t)$ is obtained by numerically integrating the differential-algebraic equation (10). Subsequently, the system's constraint forces $\boldsymbol{\Lambda}(t)$ and the actuation torque $\tau_2(t)$ are algebraically computed from the lower part of the same equation.

We now present numerical simulation results for symmetric input as in (18), with amplitude $\alpha_2 = 30°$ and frequency $\omega = 4$ [rad/s]. **Fig. 3**(a) shows time plots of the two joint angles $\phi_i(t)$. **Fig. 3**(b) shows the body velocities $\dot{x}(t), \dot{y}(t), h\dot{\theta}(t)$, and **Fig. 3**(c) shows time plots the reaction forces $\lambda_i(t)$.

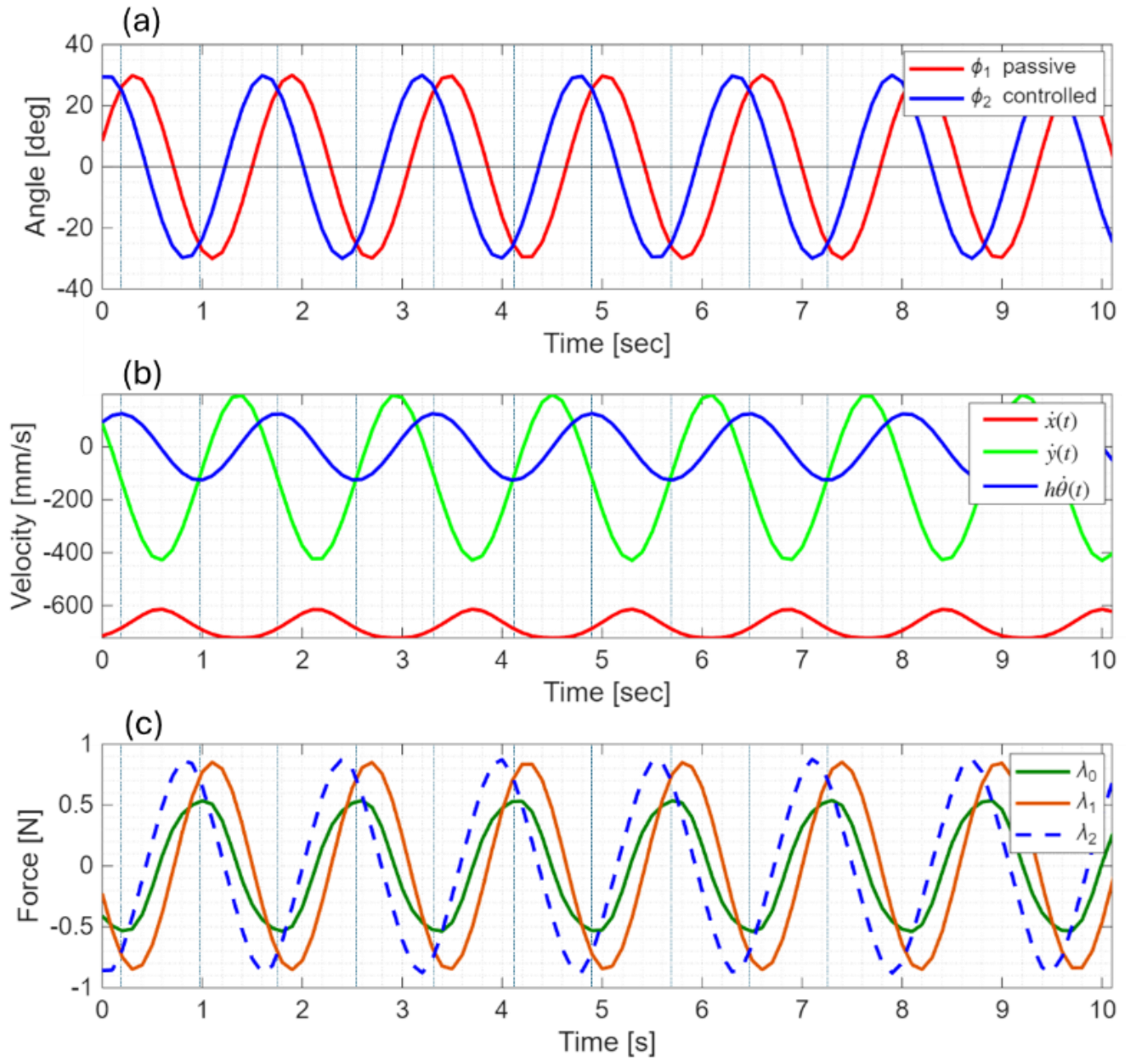

**Fig. 3**. Semi-passive configuration with symmetric input at $\omega = 4$ $[rad/s]$, after reaching steady state. (a) Joint angles $\phi_i$ vs. time. (b) Body velocities $\dot{x}, \dot{y}, h\dot{\theta}$ vs. time. (c) Reaction forces at the wheels $\lambda_i$ vs. time.

The gait trajectory plot in $(\phi_1, \phi_2)$-plane is shown in **Fig. 4** (a). It reveals a phase difference between the controlled angle and the passive angle, indicated by the elliptical shape of the steady-state curve, marked in purple. **Fig. 4**(b) shows the resulting trajectory of the robot in $x - y$ plane. Examining **Fig. 3** reveals an interesting result – even though the solution crosses singular states (where $\phi_1 = \phi_2$), the reaction forces remain bounded. The reason is that for the semi-passive configuration, the matrix **A** in (10) satisfies $\det(\mathbf{A}) \neq 0$, see [35] for details. This contrasts with the corresponding kinematic configuration where the reaction forces grow unbounded [29]. In addition, one can see from **Fig. 3**(b) that the body velocities also remain bounded. This can be explained by the fact that the resulting trajectories cross the singular line $\phi_1 = \phi_2$ at a perpendicular tangent, where the velocities $\dot{\phi}_i$ satisfy the condition (6), as can be seen in the purple elliptic loop in **Fig. 4**(a).

Next, we consider a frequency sweep for the input $\phi_2(t) = 30° \cos(\omega t)$, with a varying frequency $\omega$. Each frequency is analyzed for displacement per cycle $d$ and mean speed $\bar{v}$, which are defined above in (17). The results are presented in **Fig. 5**. A noticeable fact is the existence of two different optimum frequencies, for which displacement per cycle $d$ and the mean forward speed $\bar{v}$ attain local maxima. This important effect is corroborated below in the experimental results shown in section IV.

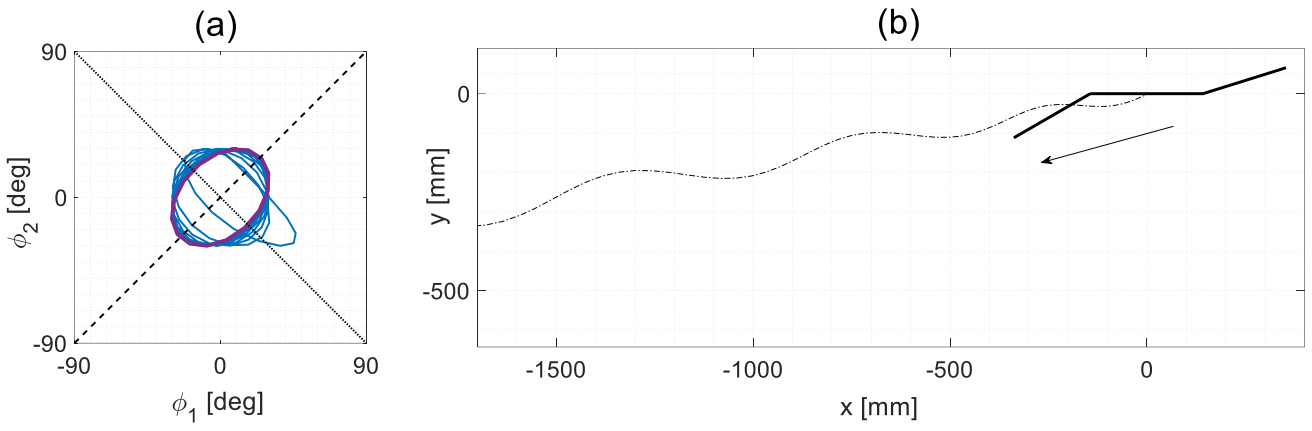

**Fig. 4**. Simulation results for the semi-passive model with a symmetric gait input at $\omega = 4$ [rad/s]. (a) Gait plot, the steady-state shape, an ellipse-like loop, is overlaid in purple. (b) x-y trajectory. The net displacement per cycle is $d$=729 [mm]. The three connected thick segments denote the robot's links at initial configuration. The dashed curve is trajectory of the middle link's center $(x, y)$, and the arrow indicates the direction of travel.

*B. Shape-actuated model, with skid and roll dissipation*

In the kinematic shape-actuated model both joints are actuated. The prescribed angle functions are given as:

$$\phi_1(t) = \gamma_1 + \alpha_1 \cos(\omega t) \qquad \phi_2(t) = \gamma_2 + \alpha_2 \sin(\omega t) \tag{19}$$

We first consider a symmetric circular gait with $\gamma_1 = \gamma_2 = 0$ and $\alpha_1 = \alpha_2 = 30°$. Under no-skid constraints, this gait is singular in its nature since it crosses the singularity condition $\phi_1 = \phi_2$. The constraint forces $\lambda_i(t)$ in such case grow unbounded sharply, in synchrony with the occurrences of singular configurations $\phi_1 = \phi_2$ [29]. Infinite reaction forces are impractical, meaning the no-skid assumption fails, at least around these specific points. Thus, we replace the no-skid constraints with frictional skid and roll resistance, as explained above in section II.

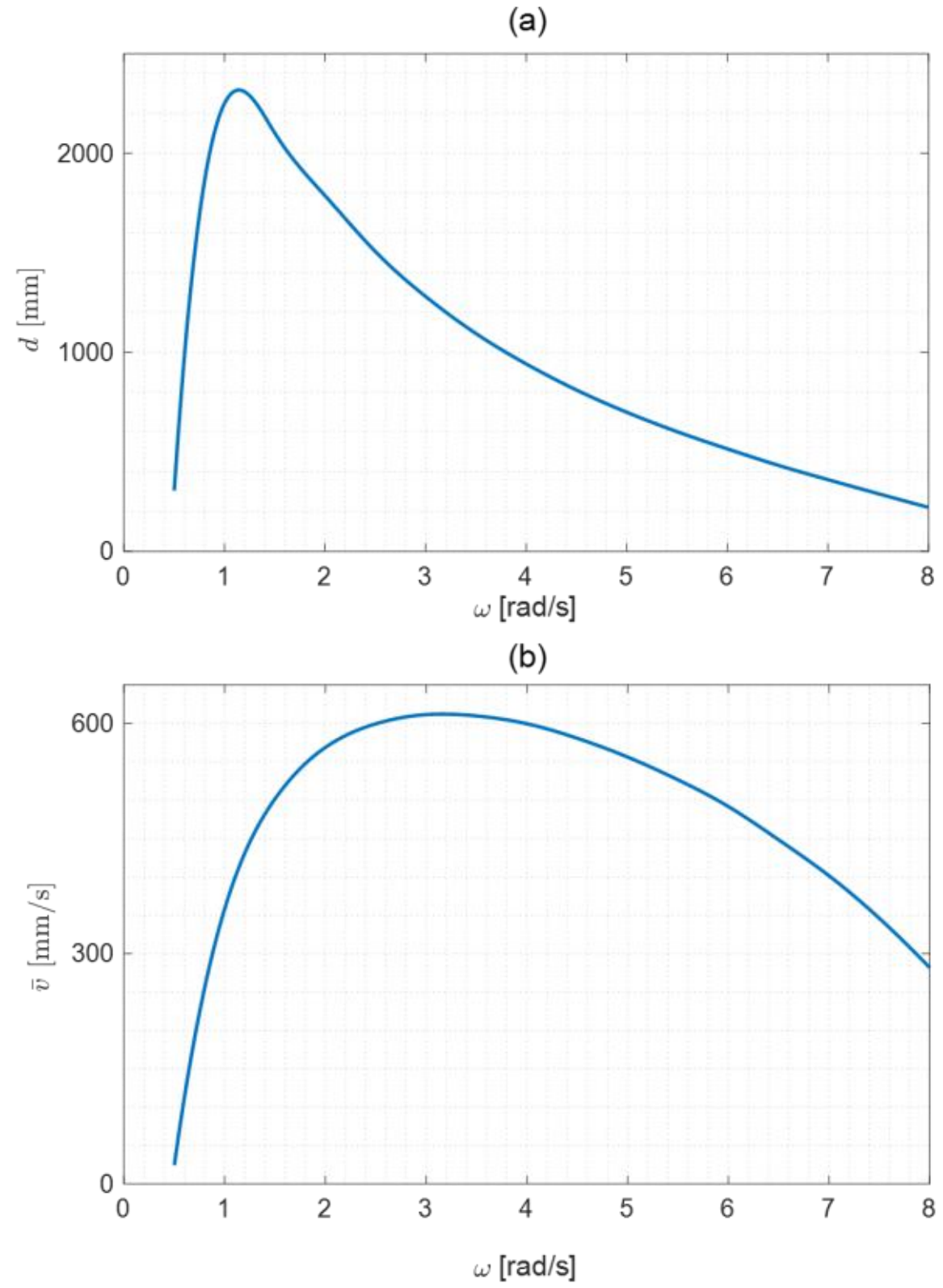

**Fig. 5.** Semi-passive configuration with symmetric input – progress parameters vs. actuation frequency. (a) Displacement per cycle $d$. (b) Mean forward speed $\bar{v}$.

The corresponding state vector is $\mathbf{z}=[\mathbf{q}_b,\dot{\mathbf{q}}_b]^T$. After introducing dissipations, the ground resistance forces (both roll and skid) are calculated directly using the viscous damping formula in (14) and (15).

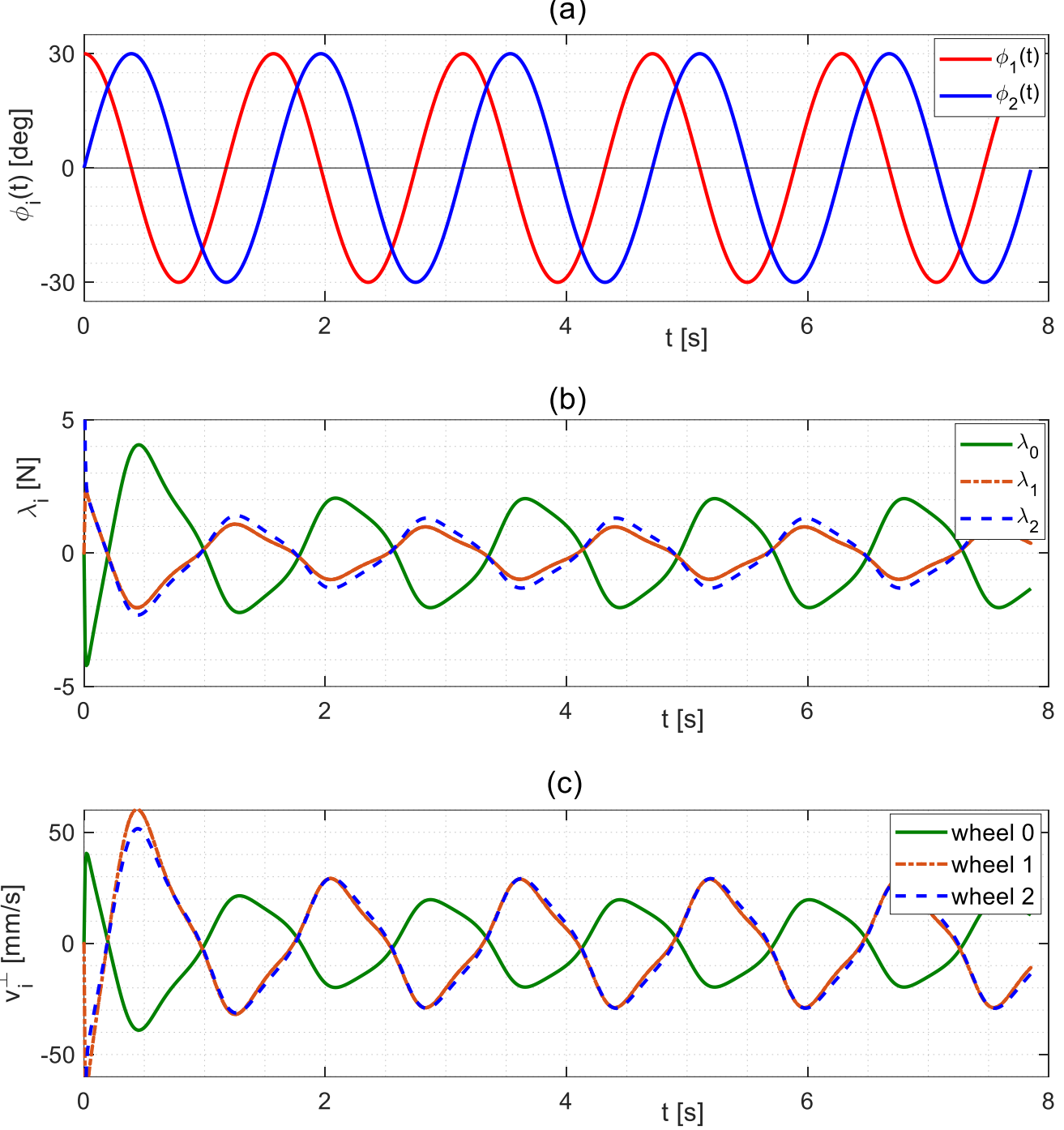

**Fig. 6.** Kinematic configuration with symmetric gait at $\omega = 4[rad/s]$. (a) Joint angles vs time. (b) Reaction forces at the wheels. (c) Skid speeds, for reference the mean forward speed is approximately 300[mm/s].

In **Fig. 6** numerical simulation results for the kinematic shape-actuated model with symmetric gait input, incorporating skid and roll dissipation, are shown. Resistance coefficients are listed in TABLE I. Notably, the inclusion of skid in the model ensures that both constraint forces and body velocities remain bounded, even when crossing singularities. This is in contrast with the basic nonholonomic case, where the reaction forces grow unbounded when crossing singularity [29]. The addition of skid dissipation to the kinematic shape-actuated model also makes the motion frequency-dependent, as will be shown in the experimental fitting results later on in **Fig. 16** and **Fig. 17** in section IV.

## IV. EXPERIMENTAL WORK AND PARAMETER FITTING

This section presents the results of the experimental work. It starts with describing the key features of the robot's mechanical design, as well as characterizing some essential parameters used in the numerical simulation. Next, results of the motion measurements for the different configurations that were tested are presented. Finally, the section ends by presenting the parameter fitting, which was performed to improve the agreement between the theoretical model and the experimental results. The results are presented following the central guideline of this study, distinguishing between a kinematic shape-actuated and a semi-passive robot.

### A. Experimental Setup

We now refer to some key features in the design of the robot, as well as describing the motion tracking setup. The robot's design and its main components are shown in **Fig. 7**. The drive system consists of two servo motors and an Arduino microcontroller. The robot is made of three links, which are assumed to be always in contact with the ground. In theory, this requires exactly three wheels, as shown in the model in **Fig. 2**. In practice, to ensure stability, the middle link is equipped with a double-wheel axle. Thus, the robot has four contact points with the ground, which no longer guarantees that all wheels are in contact with the ground. The chosen solution to this issue is the suspension unit shown in **Fig. 7**. It is designed to allow both rotational movement of the joint, as well as ensure persistent contact with the ground. This is achieved using a spring-loaded mechanism. Much similar to the role of the suspension in a car.

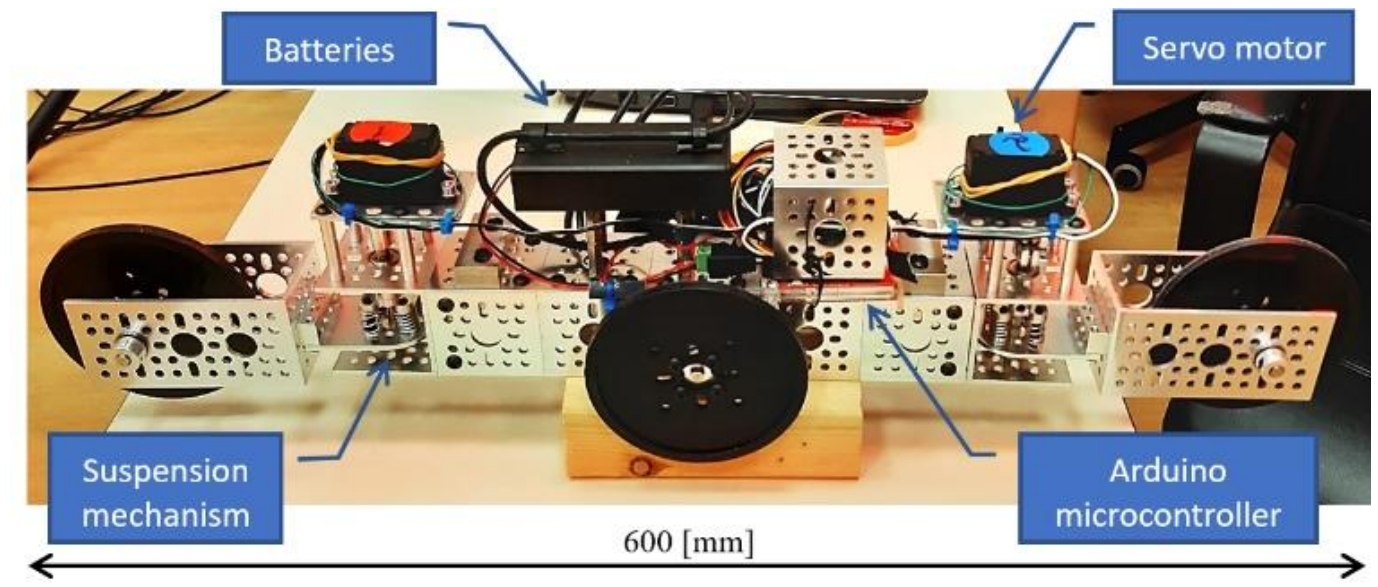

**Fig. 7.** Main componenets of the robot's mechanical design. This image shows the robot in its kinematic configuration, where both joint are controlled.

We now address the design of the passive joint and the fitting of its parameters. The passive joint in its skewed state is shown in

**Fig. 8**(a), where the free position angle is $\gamma_1 = 45°$. The rotational stiffness of the passive joint is estimated by using the stiffness of the linear spring and considering the installation geometry, assuming a linearized second-order system of inertia-spring-damper, under small-angle approximation. To verify the theoretical calculation, an experiment was carried out to measure the free angular motion of a joint that was released from a non-equilibrium state. The recorded angular displacement is shown in **Fig. 8**(b). The non-dimensional damping coefficient $\zeta$ was evaluated using the logarithmic decrement method, which was then used to calculate physical damping $c_\tau$ [35]. The parameters for a symmetric pair of springs, as shown in **Fig. 8**(c), were fitted in a similar manner. In the symmetric configuration, the free position angle is $\gamma_1=0$. The fitted values appear in TABLE II.

TABLE I presents the design parameters for the snake robot in its kinematic shape-actuated configuration, including resistance coefficients used in numerical simulations derived from experimental work, as discussed later.

TABLE II contains the relevant parameters for the robot in its semi-passive configuration. Note that the mass of link 1 increased due to addition of the spring mechanism, and a block mass has been added to link 2 in order to preserve balance and symmetry between mass properties of links 1 and 2. The reason for differences in the skid friction coefficients $c_i^S$ between the shape-actuated and semi passive configurations is that for each of these two configurations, the robot's shafts have been re-assembled, with different tuning and alignment of the suspension system and the angles between the wheels' planes and the grounds. This causes differences in the wheels friction, mainly in skid resistance. This is in addition to internal structural dissipation of the suspension mechanism. These unmodelled dissipation effects were attributed to wheels' damping in our simplified planar model

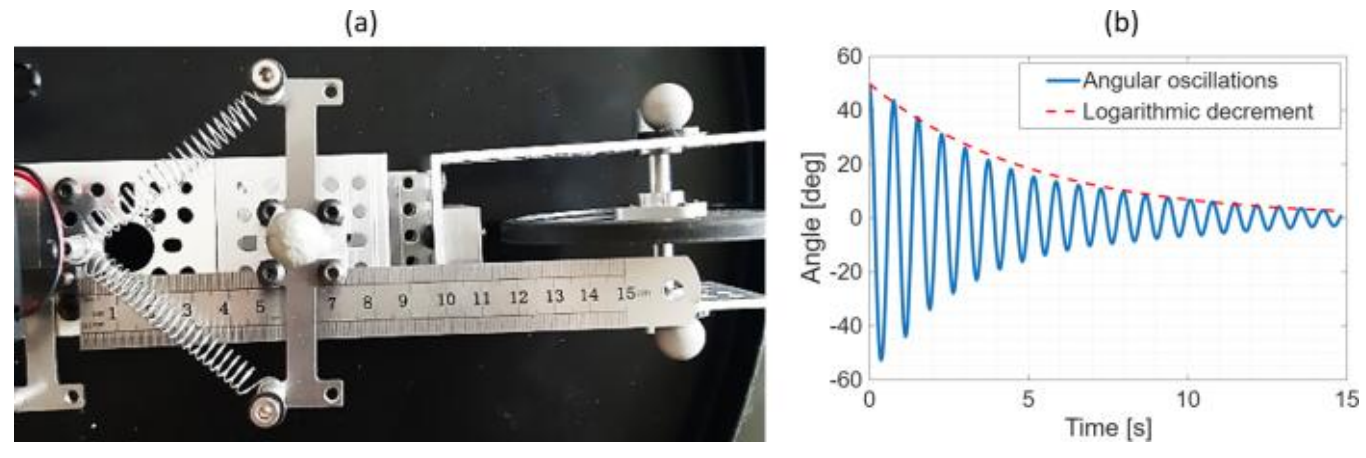

**Fig. 8.** Mechanical design of the symmetric passive joint configuration. (a) Symmetric joint $\gamma_1$=0° (b) Symmetric passive joint response to displacement from equilibrium, motion tracking and fitting of natural frequency and damping.

TABLE I

MODEL PARAMETERS FOR THE KINEMATIC SHAPE-ACTUATED ROBOT

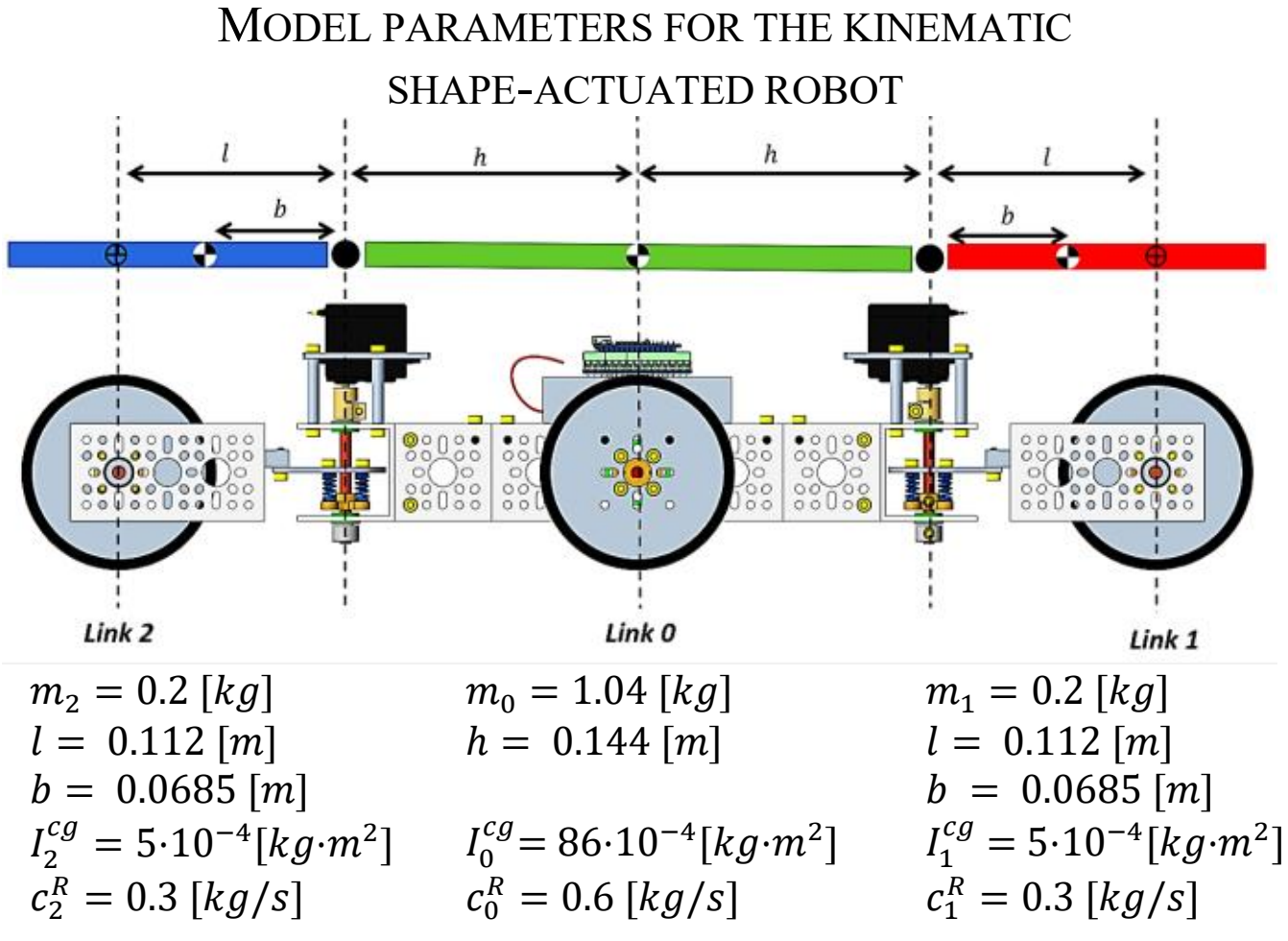

| Link 2 | Link 0 | Link 1 |
|---|---|---|
| $m_2 = 0.2\ [kg]$ | $m_0 = 1.04\ [kg]$ | $m_1 = 0.2\ [kg]$ |
| $l = 0.112\ [m]$ | $h = 0.144\ [m]$ | $l = 0.112\ [m]$ |
| $b = 0.0685\ [m]$ | | $b = 0.0685\ [m]$ |
| $I_2^{cg} = 5{\cdot}10^{-4}[kg{\cdot}m^2]$ | $I_0^{cg} = 86{\cdot}10^{-4}[kg{\cdot}m^2]$ | $I_1^{cg} = 5{\cdot}10^{-4}[kg{\cdot}m^2]$ |
| $c_2^R = 0.3\ [kg/s]$ | $c_0^R = 0.6\ [kg/s]$ | $c_1^R = 0.3\ [kg/s]$ |

**Skid resistance coefficients $c_i^S$**

| Values used in the numerical simulations | | |
|---|---|---|
| $c_2^S = 50[N{\cdot}s{\cdot}m^{-1}]$ | $c_0^S = 100[N{\cdot}s{\cdot}m^{-1}]$ | $c_1^S = 50\ [N{\cdot}s{\cdot}m^{-1}]$ |
| **Fitted values for skewed configuration** | | |
| $c_2^S = 95[N{\cdot}s{\cdot}m^{-1}]$ | $c_0^S = 74.5[N{\cdot}s{\cdot}m^{-1}]$ | $c_1^S = 91\ [N{\cdot}s{\cdot}m^{-1}]$ |
| **Fitted values for symmetric configuration** | | |
| $c_2^S = 45[N{\cdot}s{\cdot}m^{-1}]$ | $c_0^S = 104[N{\cdot}s{\cdot}m^{-1}]$ | $c_1^S = 34\ [N{\cdot}s{\cdot}m^{-1}]$ |

TABLE II

MODEL PARAMETERS FOR THE SEMI-PASSIVE ROBOT

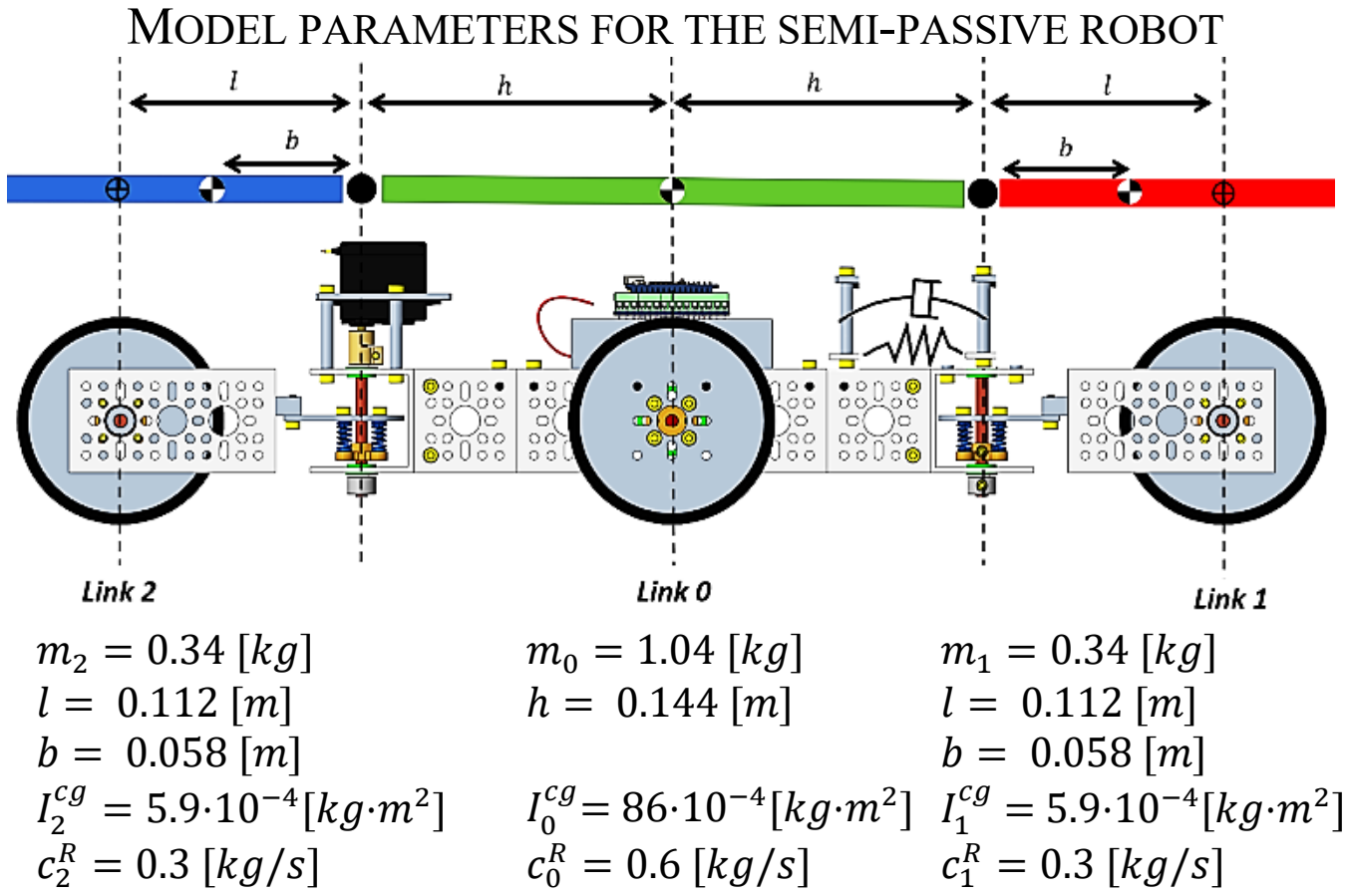

| Link 2 | Link 0 | Link 1 |
|---|---|---|
| $m_2 = 0.34\ [kg]$ | $m_0 = 1.04\ [kg]$ | $m_1 = 0.34\ [kg]$ |
| $l = 0.112\ [m]$ | $h = 0.144\ [m]$ | $l = 0.112\ [m]$ |
| $b = 0.058\ [m]$ | | $b = 0.058\ [m]$ |
| $I_2^{cg} = 5.9{\cdot}10^{-4}[kg{\cdot}m^2]$ | $I_0^{cg} = 86{\cdot}10^{-4}[kg{\cdot}m^2]$ | $I_1^{cg} = 5.9{\cdot}10^{-4}[kg{\cdot}m^2]$ |
| $c_2^R = 0.3\ [kg/s]$ | $c_0^R = 0.6\ [kg/s]$ | $c_1^R = 0.3\ [kg/s]$ |

**Skid resistance coefficients $c_i^S$ and passive joint parameters**

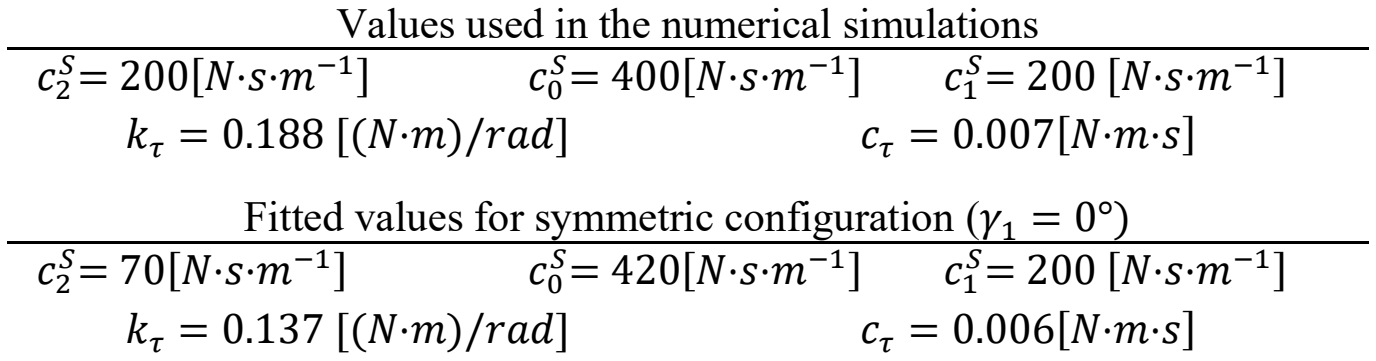

| Values used in the numerical simulations | | |
|---|---|---|
| $c_2^S = 200[N{\cdot}s{\cdot}m^{-1}]$ | $c_0^S = 400[N{\cdot}s{\cdot}m^{-1}]$ | $c_1^S = 200\ [N{\cdot}s{\cdot}m^{-1}]$ |
| $k_\tau = 0.188\ [(N{\cdot}m)/rad]$ | | $c_\tau = 0.007[N{\cdot}m{\cdot}s]$ |
| **Fitted values for symmetric configuration ($\gamma_1 = 0°$)** | | |
| $c_2^S = 70[N{\cdot}s{\cdot}m^{-1}]$ | $c_0^S = 420[N{\cdot}s{\cdot}m^{-1}]$ | $c_1^S = 200\ [N{\cdot}s{\cdot}m^{-1}]$ |
| $k_\tau = 0.137\ [(N{\cdot}m)/rad]$ | | $c_\tau = 0.006[N{\cdot}m{\cdot}s]$ |

Motion tracking was performed using the Vicon measurement system in the Biorobotics and Biomechanics Lab at the Technion. The lab is equipped with multiple cameras that surround the room, all connected to a main computer with compatible processing software. The tracked robot was marked with dedicated reflective markers at specific optimized locations. The cameras have infrared lamps used to enhance the reflectiveness of the markers. The system records the time and corresponding spatial position of each marker. The reconstruction of the robot and its kinematics from the points was done using MATLAB. Recorded at 120 Hz, the raw measurements underwent smoothing using a moving average method to reduce noise, with a window size varying from 5 to 30. The results obtained from these measurements are presented next.

### *B. Results of the kinematic shape-actuated robot*

First, we show experimental results of the kinematic shape-actuated robot with a skewed gait input. Then, we present the results of the same robot with symmetric gait input. The robot's parameters appear in TABLE I. The prescribed input for the skewed gait actuation is shown in (19), with $\gamma_1$=45°, $\gamma_2 = -\gamma_1$ and $\alpha_1 = \alpha_2$=30°. In **Fig. 9** we present the results for an actuation frequency of $\omega = 3.9[rad/s]$. As evident from **Fig. 9**(a), a slight discrepancy exists between the commanded gait input (dashed curve) and the actual measured joint angles (solid curve). A possible cause of this discrepancy is the added joint friction due to the suspension units. This effect, which alters the joint actuators' dynamics, is not accounted for in our simplified kinematic-actuation model. The actual deviation from 'perfectly symmetric' gait caused a nonzero net rotation per cycle, as depicted by the slope of angle $\theta(t)$ in **Fig. 9**(b), and clearly visible in the trajectory shown in **Fig. 9**(c). The experimental measurements, in purple curves, are compared with two numerical simulations: the blue curve is simulation under the ideal symmetric circular gait, and the green curve is simulation under an ellipse gait of the form (18) which is the best-fit ellipse of the experimental measurements. In both cases, the solution is obtained from the kinematic equation (3) of no-skid nonholonomic constraints. It can be seen that the actual curvature of the robot's trajectory as measured in the motion experiment is larger than that obtained from kinematic simulations. This indicates that dynamic effects of wheel's asymmetry may exist, as well as possible wheels' skid, which is not accounted for in (3).

Next, we conducted motion measurements across various frequencies spanning from $\omega = 0.5[rad/s]$ to $\omega = 6[rad/s]$. The mean speed $\bar{v}$ results of the robot plotted against actuation frequency, are presented in **Fig. 10** below. These findings reveal an optimal frequency, contrasting with the theoretical "purely kinematic" model's prediction of a linear relationship between speed and frequency in (3). This deviation can be primarily attributed to considerable wheel skid, as discussed later in this section.

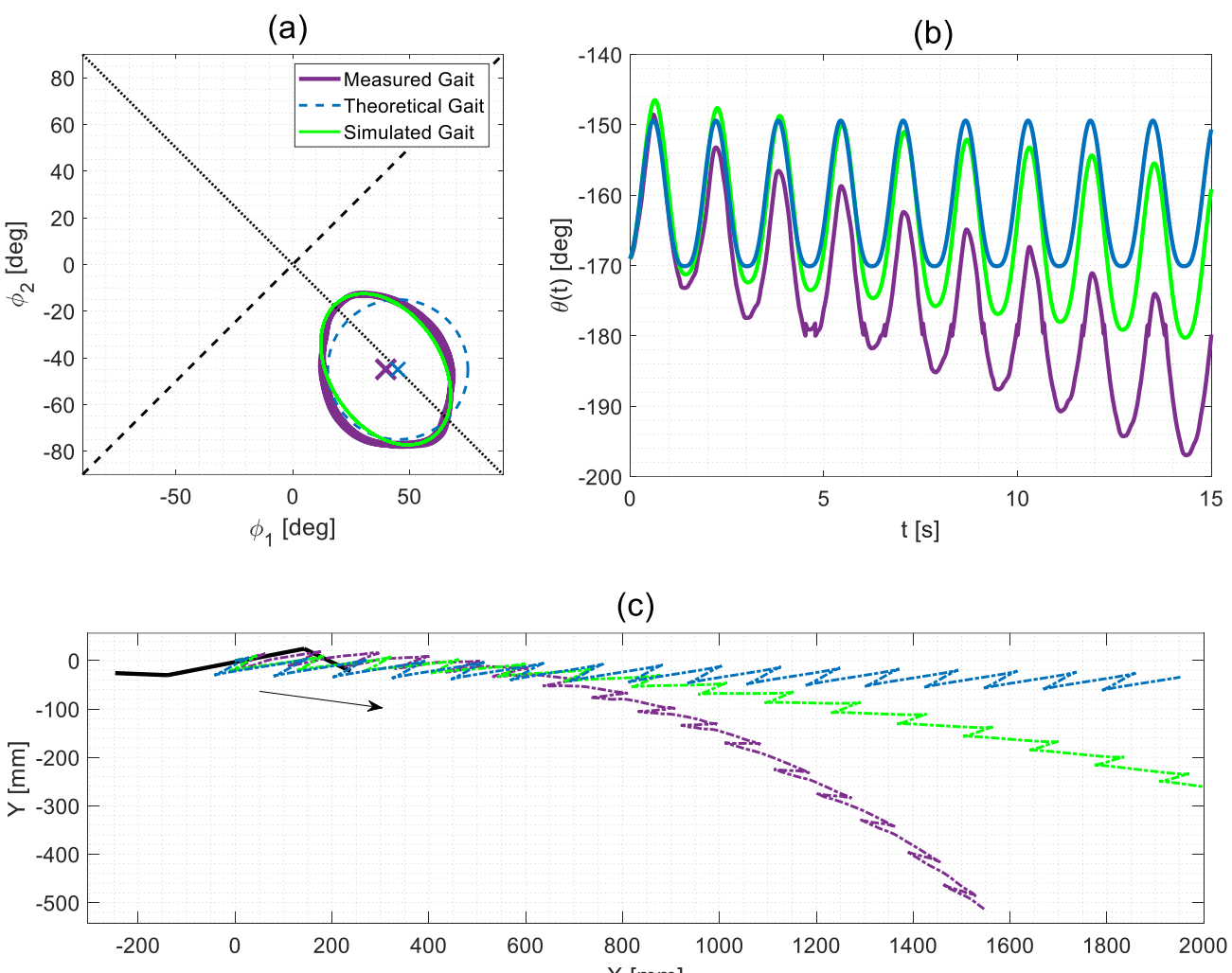

**Fig. 9.** Motion tracking results (in purple curves) of the kinematic shape-actuated robot for the skewed circular gait at $\omega$ = 3.9 [rad/s]. Simulation results are overlaid for comparison, under the ideal circular gait input (blue curves) and under the actual measured ellipse-shaped gait input (green curves). (a) Gait plot. (b) Middle link angle $\theta(t)$ plot. (c) Trajectory in the x-y plane, $d$ = 104.3 [mm]. The three connected thick segments denote the robot's links at initial state. The dashed colored curves are trajectories of the middle link's center $(x, y)$, and the arrow indicates the direction of travel.

Next, we proceed to the results for the symmetric gait input. The prescribed angular command is given in (19), where $\omega = 3.9[rad/s]$, $\gamma_1$=0°, $\gamma_2$=0°, and $\alpha_1$=$\alpha_2$=30°. Similar to the skewed input scenario, deviations of the measured gait from the ideal one are observable in **Fig. 11**(a). Once again, this results in a nonzero net rotation of the robot, evident in the trajectory plot in **Fig. 11**(c) and quantified by the slope of $\theta(t)$ in **Fig. 11**(b). In all parts of **Fig. 11**, the experimental measurements in purple curves are compared with two numerical simulations.

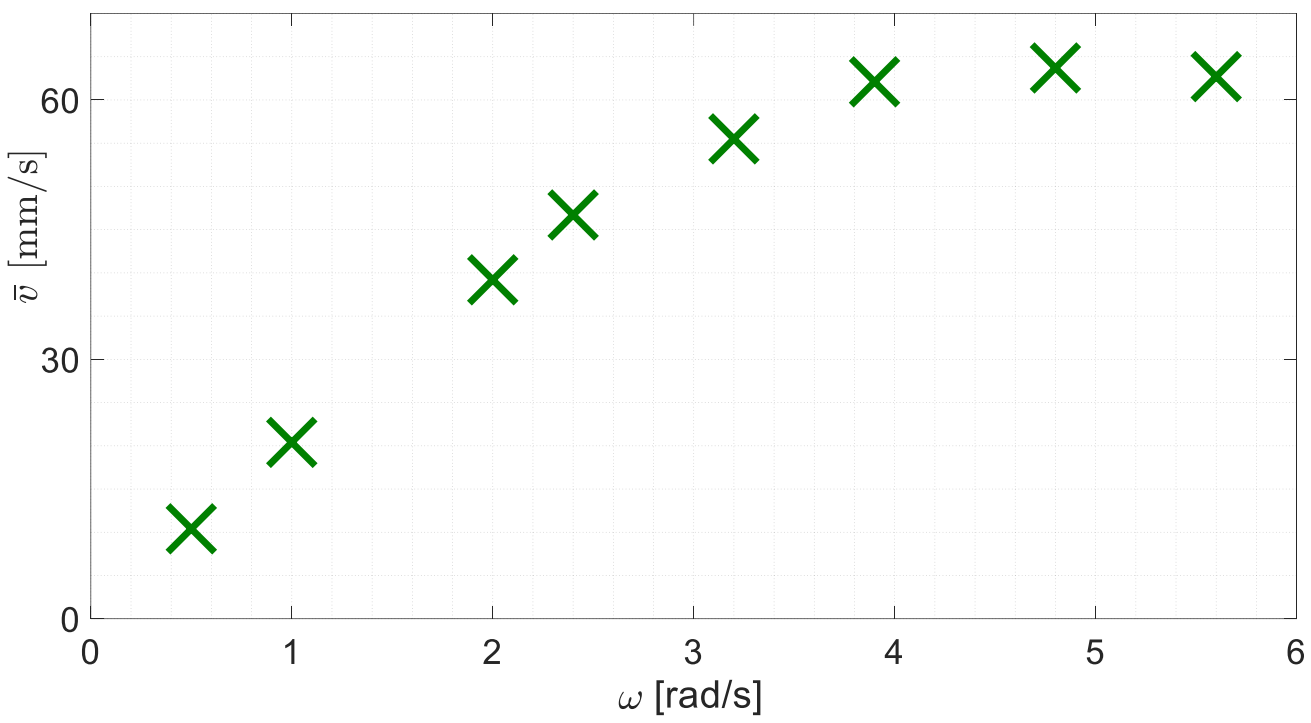

**Fig. 10.** Mean speed vs frequency for the kinematic shape-actuated robot with a skewed gait.

The blue curve is simulation under the ideal symmetric circular gait, where the solution is obtained from the kinematic equation (3). The green curve is simulation under an ellipse gait of the form (18) which is the best-fit ellipse of the experimental measurements. Since this 'imperfect' gait does not satisfy the condition (6) for crossing singularity, we use dynamic simulations with roll and skid resistance, where the skid friction coefficients $c_i^S$ are 100 times larger than the nominal values in Table I, in order to regularize the singularity crossing and attain finite reaction forces.

Measurements of the wheels' roll and skid velocities $v_i^{\parallel}(t), v_i^{\perp}(t)$ for this experiment are shown in **Fig. 12**. It can be seen that the skid velocities are significant, as expected in a symmetric gait input that crosses singular states. Some oscillation frequencies higher than the actuation frequency $\omega$ can be identified. We assume that the structure of the suspension mechanism and its inherent flexibility contribute to this phenomenon. Nevertheless, this aspect was not investigated further in the current study.

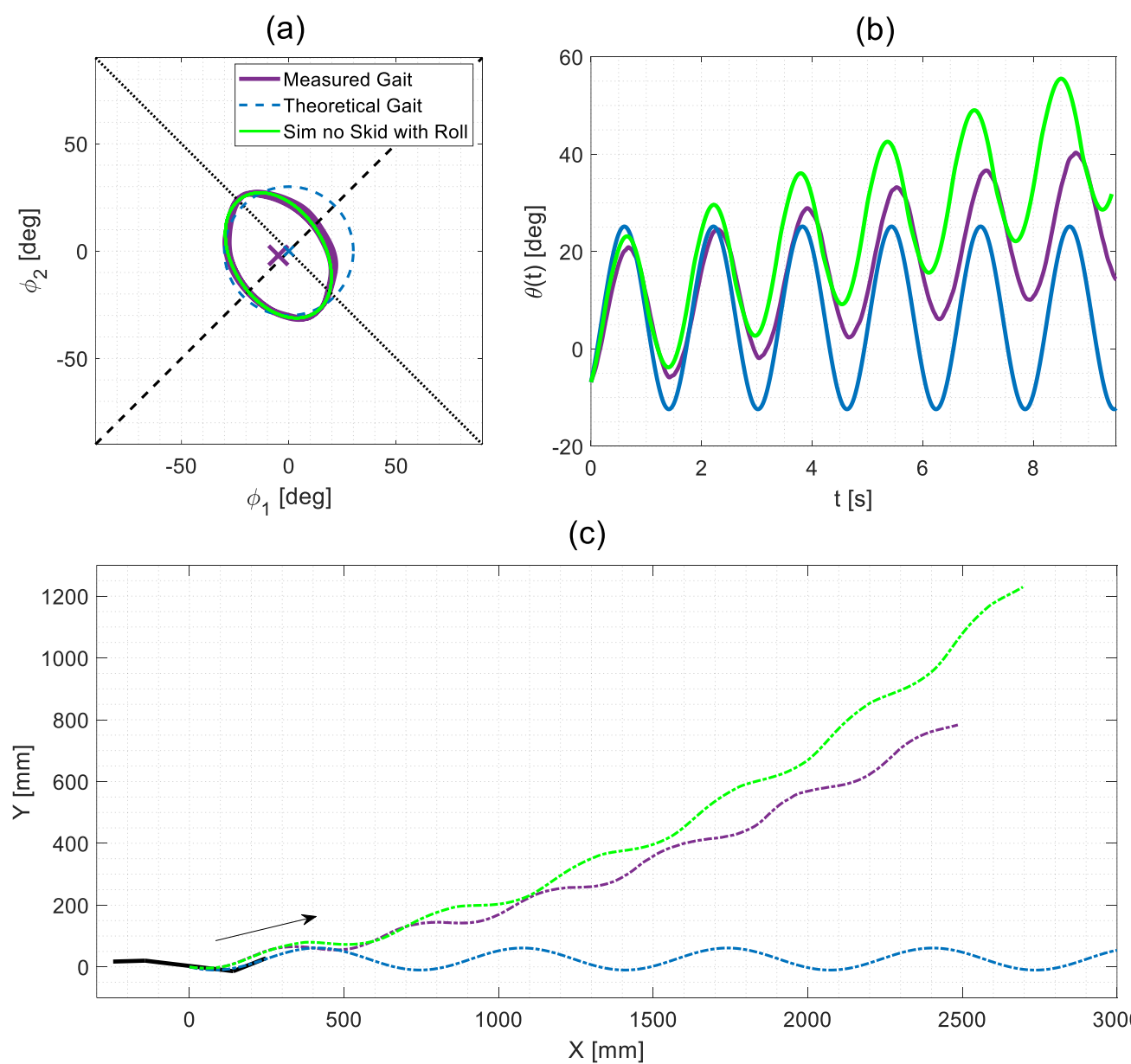

**Fig. 11.** Motion tracking results for the kinematic shape-actuated robot, symmetric gait at $\omega$ = 3.9 [rad/s]. (a) Gait plot. (b) Middle link $\theta(t)$ plot. (c) Trajectory in the x-y plane, $d$ = 437 [mm].

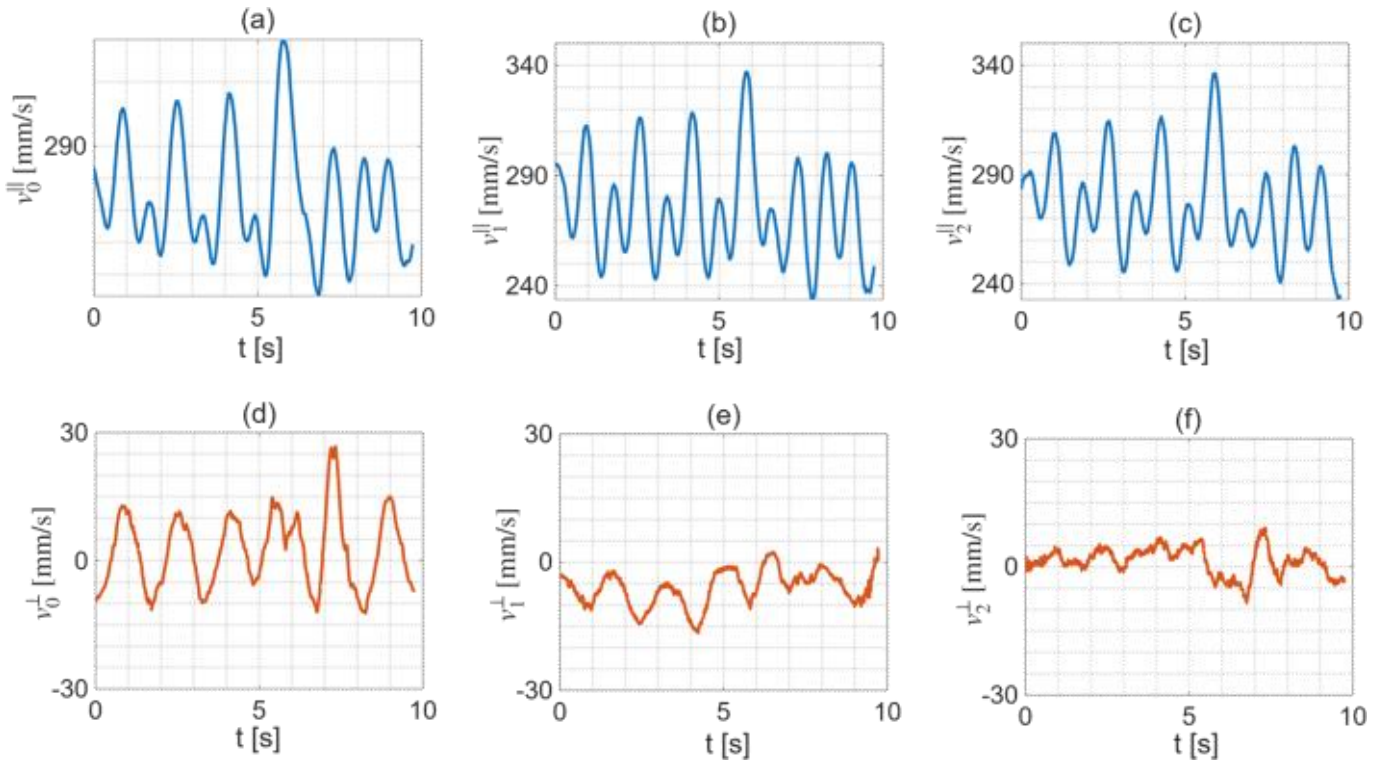

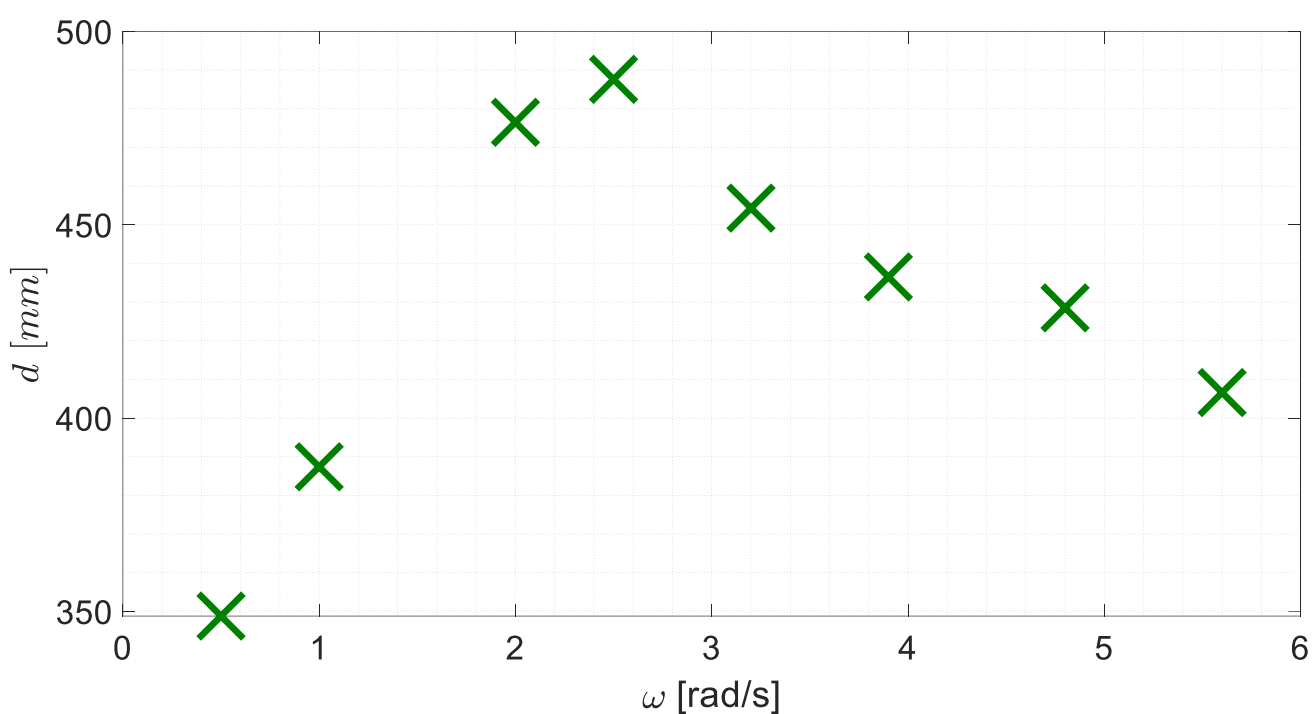

**Fig. 12.** Measurements of wheels' roll and skid velocities for each wheel for the kinematic shape-actuated robot under symmetric gait with $\omega$ = 3.9 [rad/s].

As before, the motion measurements for the symmetric actuation were conducted across various frequencies. In **Fig. 13** the displacement per cycle $d$ results are plotted against actuation frequency. The results exhibit an optimum at around $\omega$=2.5[rad/s]. The basic nonholonomic, kinematic shape-actuated model has no frequency dependence, meaning that in this model, the distance per cycle $d$ does not change with frequency. Numerical simulations of the kinematic shape-actuated model, incorporating wheel skid, exhibit qualitative agreement with these findings, see **Fig. 17** below. Therefore, this phenomenon is attributed to skid of the wheels.

**Fig. 13.** Displacement per cycle $d$ vs frequency for the kinematic shape-actuated robot with a symmetric gait

## C. Results of the semi-passive robot

We now present the results of the semi-passive robot with a symmetric gait input, which are presented in **Fig. 14**. The input command for $\phi_2(t)$ is given in (18), with $\alpha_2 = 30°$. The free angle of the passive joint is set to $\gamma_1 = 0$, as shown in **Fig. 8**(a). The purple curve in **Fig. 14** is the gait trajectory obtained from experimental measurements. The blue curve is obtained in steady-state from numerical simulations under the same actuation input, assuming no-skid constraints and zero rolling resistance. One can see that the aspect ratio of the obtained ellipse differs significantly from the experimentally measured gaits, which indicates a large difference in the phase shift between the actuated and passive joint angles. In contrast, the green curve is obtained in steady-state from numerical simulations under the same actuation input, assuming no-skid constraints and incorporating rolling resistance. One can see that this resulting gait is much closer to the experimental measurements. This indicates the importance of incorporating the wheels' rolling resistance in the dynamic model, in addition to the passive joint's damping.

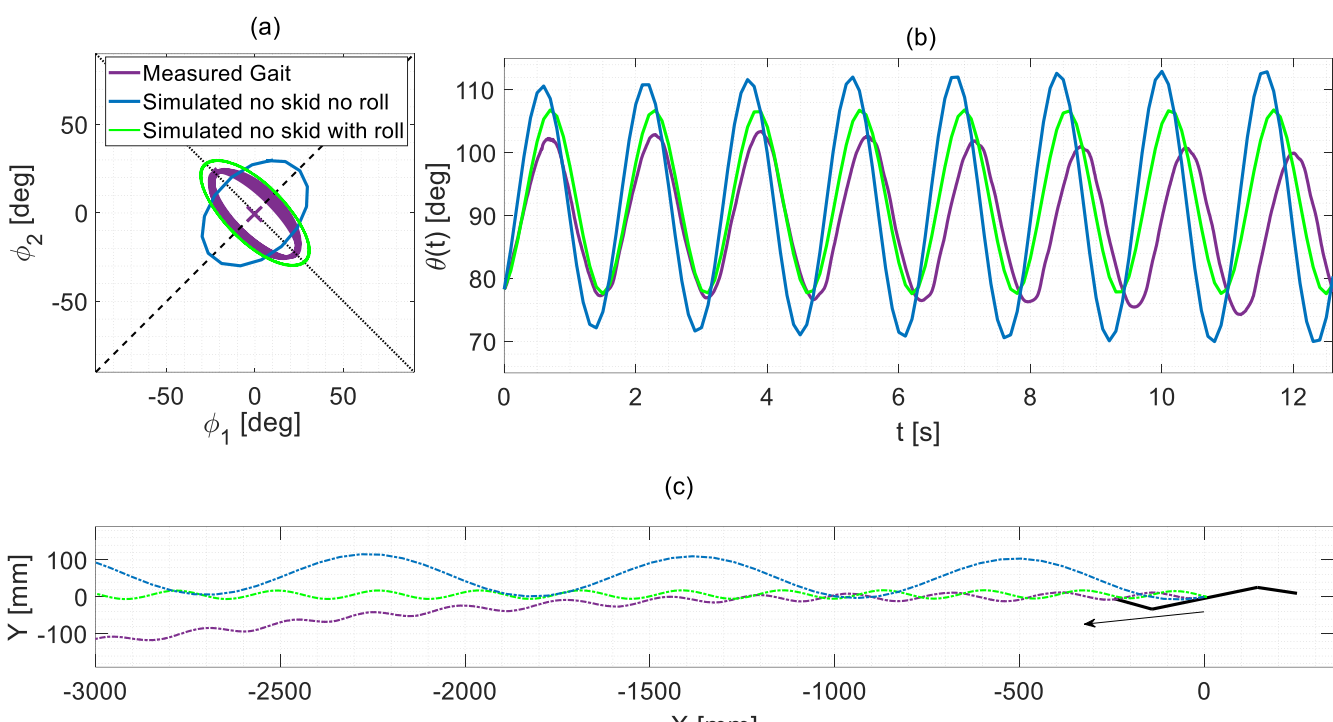

**Fig. 14.** Motion tracking results for the semi-passive robot, symmetric gait at $\omega$ = 3.9 [rad/s]. (a) Gait plot. (b) Middle link $\theta(t)$ plot. (c) Trajectory in the x-y plane, $d$ = 263 [mm].

A noteworthy result is depicted in **Fig. 15**, where the mean speed reaches an optimum at around $\omega = 4\ [rad/s]$. Interestingly, a similar optimum has been predicted in the numerical simulation (**Fig. 5**(b)) and will be addressed later in **Fig. 18**(b).

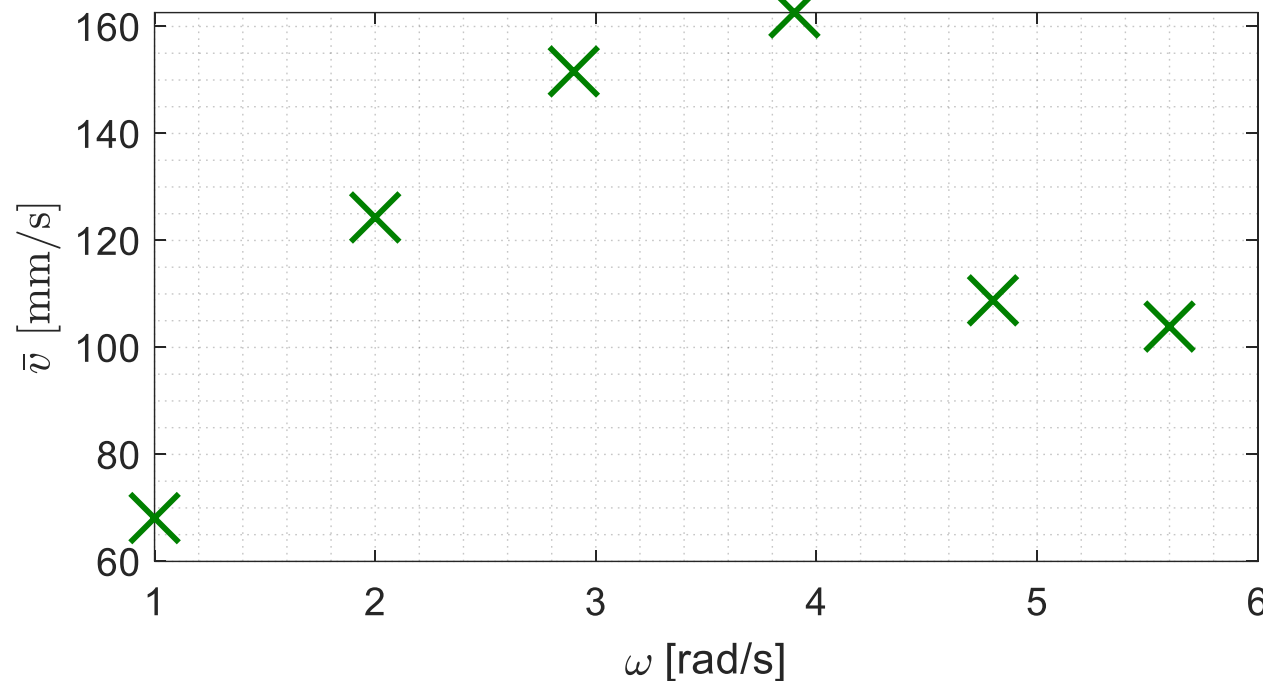

**Fig. 15.** Mean speed $\bar{v}$ vs. actuation frequency $\omega$ for the semi-passive robot with a symmetric input.

## D. Parameter Fitting of the Theoretical Models

We now present the parameter fitting process, which is performed in order to improve the agreement between the theoretical models and the experimental results. A primary goal is to calibrate the various energy loss mechanisms discussed in section II. The fitting is done for each of the two main configurations of the robot: kinematic and semi-passive actuation. To evaluate and determine the best fit, several theoretical models were tested for each type and compared, focusing primarily on criteria such as net displacement and mean speed. This section concludes by presenting a comparison of the skid values among the various configurations, both in relation to each other, and in comparison to the results obtained from the numerical simulation.

As explained in section II, energy dissipation mechanisms have been added to the theoretical models in order to make them more realistic. The kinematic model has six such parameters:

the skid resistance coefficients $c_i^S$ and the roll resistance coefficients $c_i^R$, for $i = 0,1,2$. The semi-passive model has an additional damping coefficient $c_\tau$ for the compliant joint – making it a total of seven parameters. The rolling resistance coefficients $c_i^R$ was estimated from an independent experiment of pushing the robot with its joints fixed at straightened configuration $\phi_1 = \phi_2 = 0$, measuring the rate of decay of the forward speed, and fitting to linear damping, see Fig. 55 in [35]. The passive joint's damping coefficient $c_\tau$ was also estimated from independent experiment, see **Fig. 8**(b) and explanation above. We now present the parameter fitting performed for the skid resistance coefficients. The parameter fitting was conducted using MATLAB function '*fmincon'*. It is used to find the minimum of a constrained nonlinear multivariable function. In this case, the objective function performs a numerical simulation and returns a scalar error value, which represents the difference between the simulation and the experimental results. The value of the error depends on several distinctive metrics, each of them having a specific weight assigned to it. It is worth noting that normalizing the variables is essential before assigning weights to them, in order to prevent any bias resulting from differences in orders of magnitude.

The chosen metrics for the objective function are displacement per cycle $d$, mean speed $\bar{v}$, skid ratio for each wheel $\sigma_i$, and the oscillation amplitude of the passive joint $\alpha_1$ (only for the semi-passive configuration). We define the skid ratio as a comparative quantitative value which represents a measure for skidding during the robot's periodic motion. The skid ratio is defined in (20) below.

$$\sigma_i = \frac{\sqrt{\frac{1}{N}\sum_{j=1}^{N}\left(v_i^\perp(t_j)\right)^2}}{l\omega} \tag{20}$$

For $N$ discretized times $t_j$ within a period. This formula represents the ratio between the RMS value of the perpendicular speed of the wheel $v_i^\perp$, to the characteristic speed of the side arms $l\omega$.

We now define the deviation variables to be used in the objective function. Each deviation variable represents the absolute value of the difference between the experimental and simulation results (upon reaching a steady state). To ensure consistent scaling across all parameters, these deviation variables are normalized by the experimental values.

$$e_d = \left|\frac{d_{exp} - d_{sim}}{d_{exp}}\right| \qquad e_v = \left|\frac{\bar{v}_{exp} - \bar{v}_{sim}}{\bar{v}_{exp}}\right|$$
$$e_i^\sigma = \left|\frac{\sigma_i^{exp} - \sigma_i^{sim}}{\sigma_i^{exp}}\right| \qquad e_\alpha = \left|\frac{\alpha_{exp} - \alpha_{sim}}{\alpha_{exp}}\right| \tag{21}$$

The objective function to be minimized is then defined as

$$J = w_d e_d + w_v e_v + w_\sigma(e_1^\sigma + e_2^\sigma + e_3^\sigma) + w_\alpha e_\alpha \tag{22}$$

The weights, denoted as $w_i$, were determined through a manual adjustment of the weight allocation and assessing the comparison results for each case. The chosen weights for the different configurations are listed in TABLE III. Overall, a higher weight was assigned to displacement per cycle compared to other metrics, as it consistently resulted in the most optimal fit. In the semi-passive configuration, emphasis was placed on the amplitude of the passive joint, due to its crucial role in dynamic response.

TABLE III

PARAMETER FITTING WEIGHTS FOR DIFFERENT CONFIGURATIONS

| | $w_d$ | $w_v$ | $w_\sigma$ | $w_\alpha$ |
|---|---|---|---|---|
| Kinematic with skewed gait | 10 | 1 | 1 | - |
| Kinematic with symmetric gait | 10 | 1 | 0.2 | - |
| Semi-passive with symmetric gait | 10 | 1 | 1 | 50 |

**Fig. 16** displays a comparison between the experimental results and several theoretical models, for the kinematically-actuated case with skewed input. The results clearly indicate that assuming no-skid constraints is insufficient in this case, and the incorporation of skid into the model is essential for achieving a satisfactory qualitative agreement with the experimental findings. The standard no-skid theoretical model predicts no dependency of the displacement per cycle on actuation frequency for the kinematic shape-actuated case. Therefore, the experimental results can only be explained by adding skid dissipation. The blue circles in **Fig. 16** represent the numerical simulation results of the no-skid model using the actual gait inputs, as obtained from the experimental measurements (mean values and oscillation amplitudes of both actuated joints' angles). These simulation results aim to demonstrate that solely employing the precise input gait values in the no-skid model cannot explain the observed discrepancies.

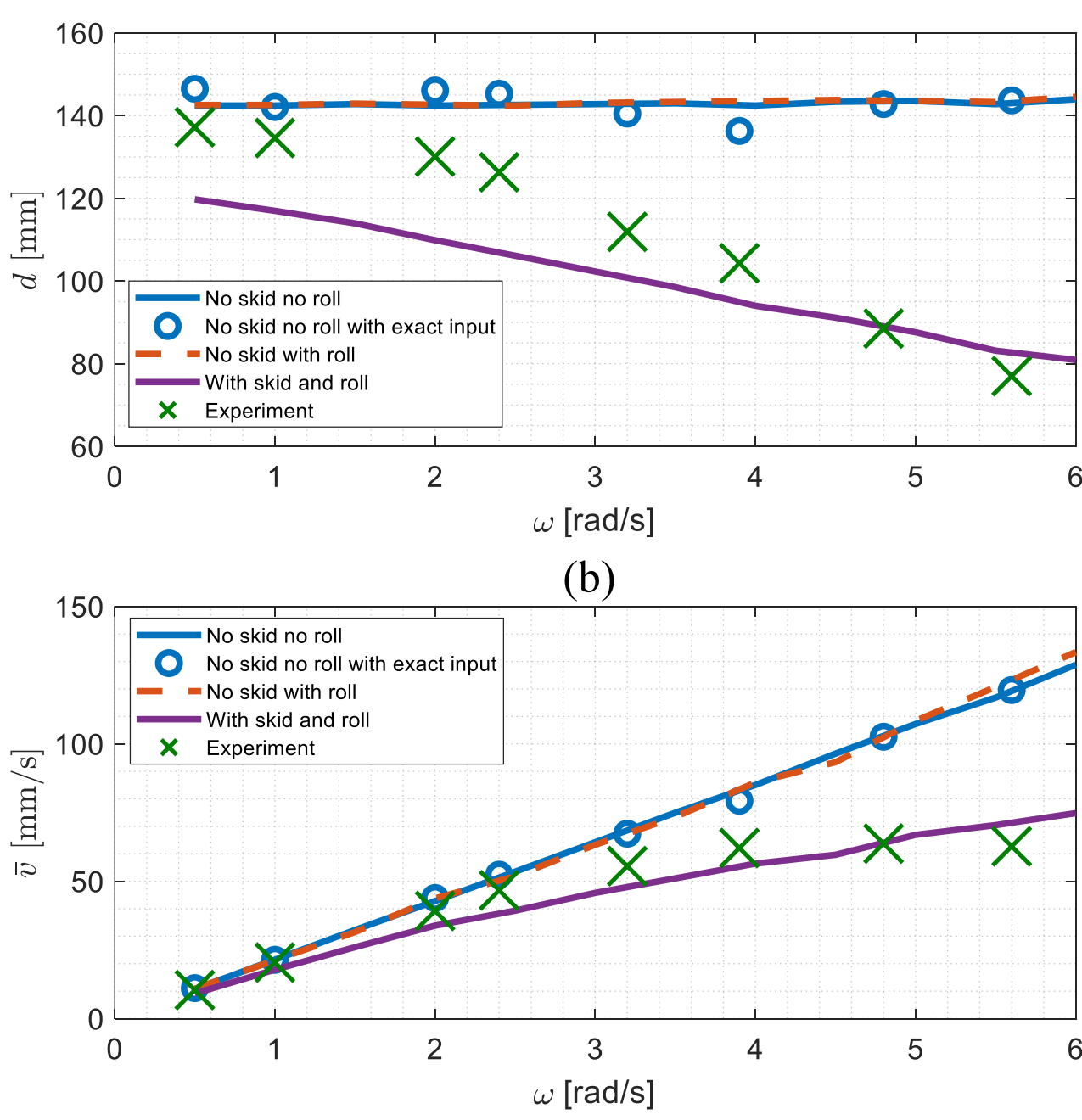

**Fig. 16.** Comparison with theoretical models - kinematic shape-actuated robot with skewed gait. (a) Displacement per cycle vs actuation frequency. (b) Mean speed vs actuation frequency.

**Fig. 17** presents a comparison between the experimental results for the symmetric gait input and three theoretical models. In this case, simulating the actual measured joint angles as inputs to the no-skid model is not feasible. This is because simulating a non-circular input gait introduces a singularity crossing, resulting in the divergence of the solution and preventing numerical integration, as explained in [29]. In any case, as seen in the skewed gait, this alone would not explain the difference between the theoretical model and the experimental results. When comparing the symmetric and skewed gaits, the symmetric gait exhibits considerably greater distance per cycle values. Looking at **Fig. 17**, a substantial difference is observed between the simulations with nonholonomic no-skid constraints and those incorporating skid. The inclusion of skid dissipation significantly improves the agreement between the model and experimental results. The distance per cycle in the experimental results reaches an optimum around an actuation frequency of ω=2.5 [rad/s].

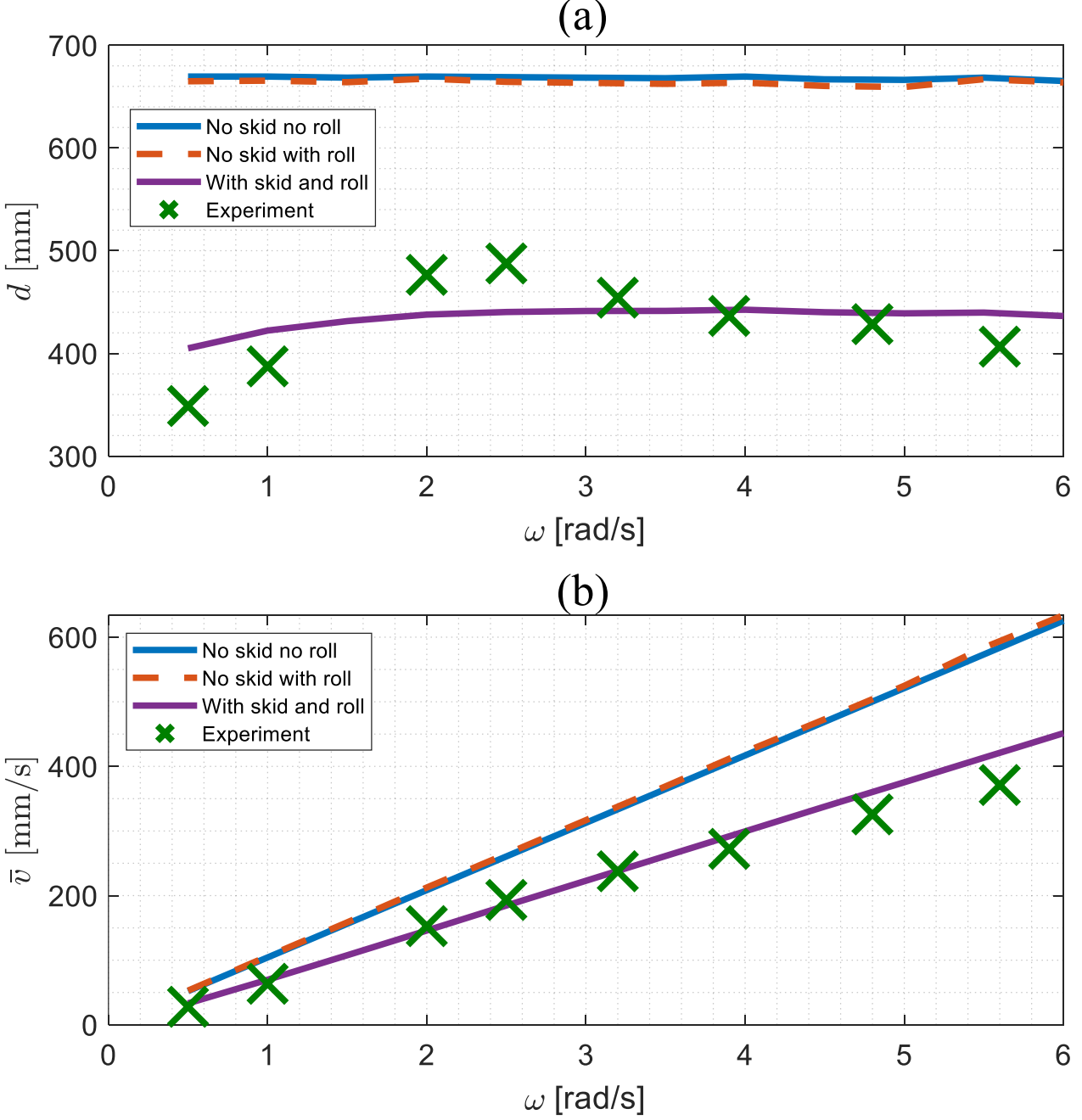

**Fig. 17.** Comparison with theoretical models - kinematic shape-actuated robot with symmetric gait. (a) Displacement per cycle vs actuation frequency. (b) Mean speed vs actuation frequency.

The parameter fitting for the semi-passive robot configuration was only done for the symmetric gait actuation. In relation to the skewed gaits, there have been difficulties in achieving sufficient progress within the desired frequency range, leading to limited data collection. Additionally, significant variations in skid ratios between test runs have posed challenges in performing a fitting over the measured range.

For the semi-passive configuration, an additional theoretical model was added to the comparison. A model that uses no-skid constraints but aims to compensate for the energy loss by adjusting the damping coefficient of the rotary spring of the compliant joint. That is, $c_\tau$ has been optimized in the fitting process, rather than taken from the independent experiment (as explained in part A of this section).

Upon examining **Fig. 18**, it becomes apparent that the standard, no-skid nonholonomic model, represented by the blue curve, which does not include any additional energy dissipation mechanisms, deviates the most from the experimental results. The results suggest that all three models, which account for additional energy dissipation, offer reasonably accurate approximations. Nevertheless, incorporating wheels' skid is unnecessary for obtaining good agreement.

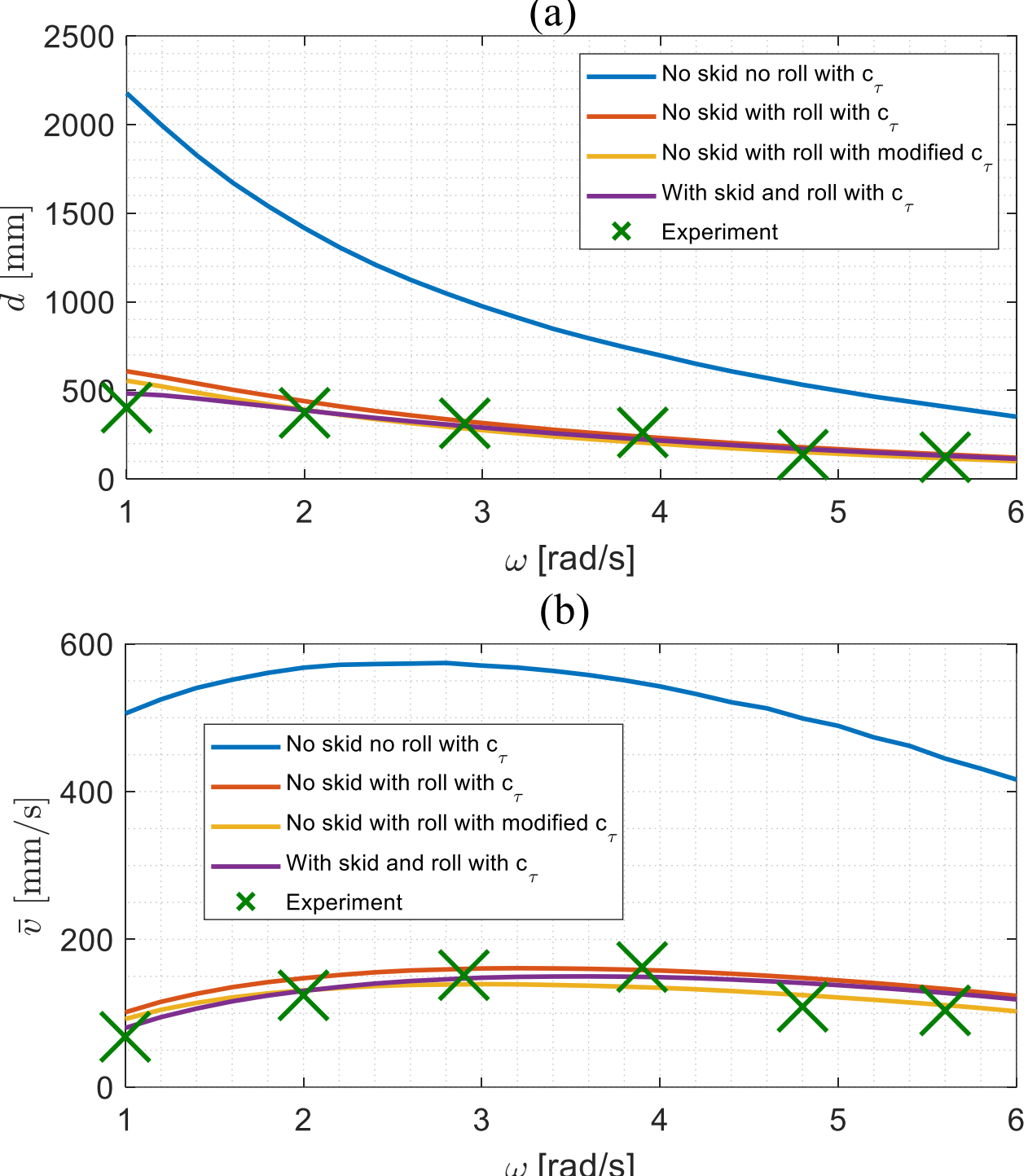

**Fig. 18.** Comparison with 4 theoretical models – semi-passive robot with symmetric gait. (a) Displacement per cycle vs actuation frequency. (b) Mean speed vs actuation frequency.

Next, we compare the skid results obtained from the different tested configurations and gaits among themselves, as well as against the results of numerical simulations. The skid values were calculated using the method defined in (20).

**Fig. 19** presents a skid comparison between symmetric and skewed gaits for the kinematic shape-actuated robot. The numerical simulations were performed using the kinematic shape-actuated model with the fitted parameters. Overall, it is evident that the skid values in the symmetric gait are greater than those in the skewed gait. This difference can be attributed to the singularity crossing that occurs during the symmetric gait actuation, which facilitates skidding.

**Fig. 20** presents a comparison of skid between the semi-passive and kinematic shape-actuated robots, both with symmetric gait input. The figure includes both experimental and numerical simulation results. The simulations were carried out using kinematic and semi-passive models with fitted parameters. It can be concluded that the semi-passive robot exhibits lower skid values compared to the kinematic robot. This difference is attributed to the dynamic behavior of the semi-passive robot, which enables it to cross singularities with bounded constraint forces (since $\det(\mathbf{A}) \neq 0$), as already observed in **Fig. 3**. This observation also provides insight into the fitting results shown in **Fig. 18**. Since the semi-passive robot experiences less skidding, it can be fitted more effectively using models with nonholonomic, no-skid constraints.

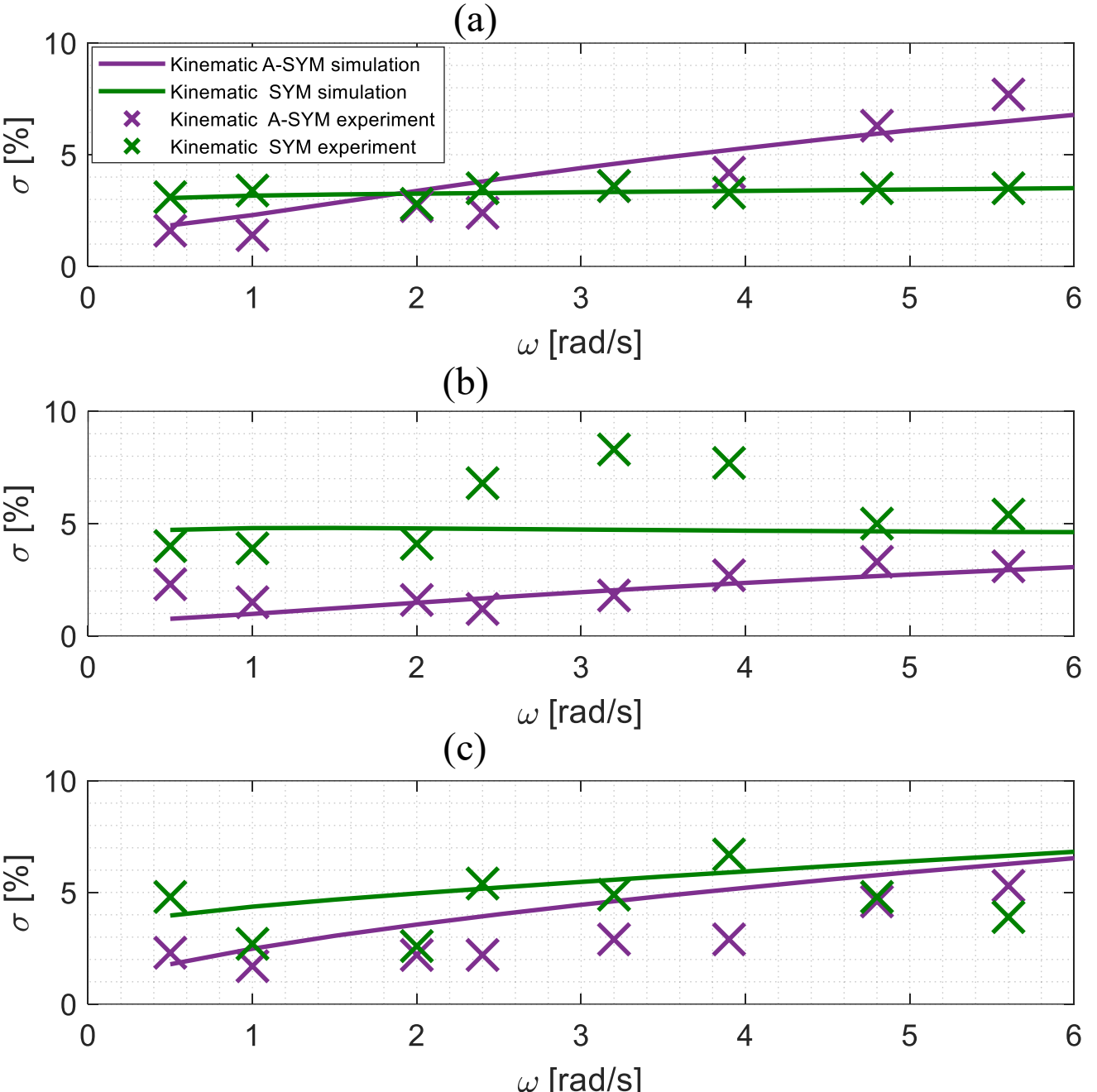

**Fig. 19.** Skid comparison - kinematic shape-actuated robot. (a) Skid ratio for wheel 0. (b) Skid ratio for wheel 1. (c) Skid ratio for wheel 2.

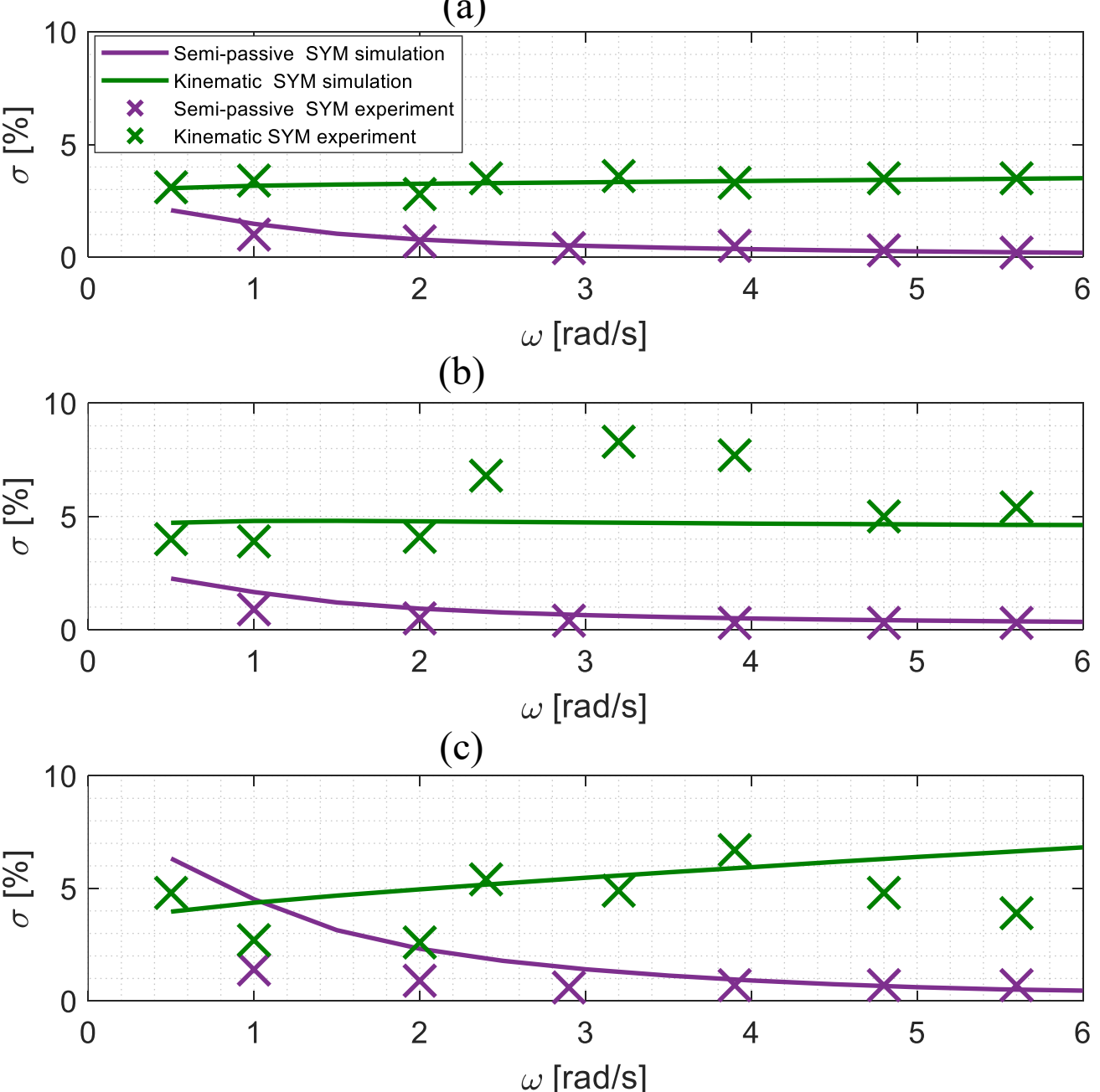

**Fig. 20.** Skid comparison - kinematic vs. semi-passive configuration for symmetric input. (a) Skid ratio for wheel 0. (b) Skid ratio for wheel 1. (c) Skid ratio for wheel 2.

## V. Concluding Discussion

This work investigated the locomotion dynamics of an underactuated three-link robotic vehicle. It focused on analyzing the configurations of fully kinematic and semi-passive actuation, while considering cases with and without singularity crossing. As part of this work, motion tracking experiments were conducted using a wheeled three-link robot that was specifically designed for this purpose. The motion of the robot was analyzed for both configurations, kinematic and semi-passive, and for different actuation inputs. The results revealed significant lateral skid, in contrast to the no-skid assumption used in standard nonholonomic models. Therefore, modified dynamic models were proposed, incorporating wheels' skid resisted by viscous friction in ground reaction forces, as well as rolling resistance. Additionally, parameter fitting was conducted in order to improve the agreement between the theoretical models and the experimental results by adjusting the viscous damping coefficients.

The primary findings concerning the kinematic configuration are as follows. The motion tracking results indicated non-trivial frequency dependence which can only be explained by adding skid dissipation to the model. This addition of skid dissipation enables the kinematic model to pass singularities while the reaction forces remain bounded (in contrast to the common no-skid nonholonomic model).

Regarding the semi-passive configuration, the main findings can be summarized as follows. The semi-passive model exhibits a resonance-like maximum in oscillation amplitude of the passive joint angle as a function of frequency, similar to frequency response of a second-order damped mechanical oscillator. Consequently, the model exhibits optimal actuation frequency for mean forward speed. This behavior in numerical simulations is corroborated in the experiments. The motion of a semi-passive robot can be reasonably predicted even with a model containing no-skid constraints, provided that rolling resistance and compliant joint's damping are considered. When skid dissipation is added to the semi-passive robot's model, it improves its accuracy in predicting the passive joint's angle amplitude, especially at high frequencies. Furthermore, the semi-passive model can cross singularities while the reaction forces remain bounded, even when using no-skid nonholonomic constraints, due to inertial effects [29,38].

Naturally, there were certain limitations to this study, briefly discussed as follows. First, the experimental aspect can be improved by redesigning the suspension mechanism to mitigate its current drawbacks, which mainly result in adding off-plane motion. Furthermore, experiments of the semi-passive robot under skewed gait inputs could be improved to get more accurate measurements.

Second, for the theoretical model, incorporating Coulomb-type friction [29] into the numerical simulations in relevant scenarios may enhance their fidelity. A more rigorous analysis of experimental measurements in attempt to devise a more realistic ground friction model for both skid and roll is an open problem that requires further exploration. Another possible direction for future extension of the model is accounting for joints' friction and its effect on actuators' dynamics. Referring to the semi-passive simulations, stability and uniqueness of the periodic solution have been achieved by utilizing realistic damping values. Using lower damping values, which may not be attainable in a realistic physical robot, can lead to the discovery of multiple periodic solutions, stability transitions, and bifurcations. This phenomenon requires further extensive analysis, presenting an ongoing challenge for future research.

Finally, regarding the theoretical analysis, several intriguing research questions remain open. For instance, exploring why is it easier for the semi-passive model to cross singularity while the velocities and forces remain bounded. A similar observation was made in [29], where the model was driven into a singular configuration by torque input, and exhibited bounded reaction forces and velocities, suggesting that the kinematic singularity may be recoverable. In fact, the work [38] argued that gaits passing through the singularity are often significantly more energy-efficient. Another question to investigate is why the symmetric gait achieves faster movement compared to the skewed gait. Although a partial explanation can be found in [15], which shows that a three-link micro-swimmer achieves greater net translation with symmetric gait input, it remains an intuitive and partial explanation, particularly since the micro-swimmer's dynamics does not have singularities.

In summary, we have shown that in analyzing mathematical models of underactuated wheeled robot locomotion, one should consider extending beyond the commonly used paradigms of kinematic actuation and no-skid nonholonomic constraints. This enables achieving a more realistic description of the robot's dynamic behavior as obtained in experiments.

## VI. Acknowledgment

The authors wish to convey thanks to Professor Alon Wolf and the BRML lab team at the Technion for their assistance and for allowing us to utilize their laboratory facilities for conducting experiments and motion measurements. In addition, sincere appreciation is extended to Ari Dantus for his substantial contributions in mechanical design of the robot prototypes, executing the motion tracking experiments, and analyzing their results. Finally, we wish to deeply thank the anonymous reviewers for their important and insightful comments, which helped us in revising this paper.